\documentclass{article}

\usepackage{arxiv}
\usepackage[utf8]{inputenc} 
\usepackage[T1]{fontenc}    
\usepackage{hyperref}       
\usepackage{url}            
\usepackage{booktabs}       
\usepackage{amsfonts}       
\usepackage{nicefrac}       
\usepackage{microtype}      
\usepackage{lipsum}		
\usepackage{graphicx}
\usepackage[numbers]{natbib}
\usepackage{doi}
\usepackage{amsmath}
\usepackage[capitalize]{cleveref}
\usepackage{multirow}
\usepackage{colortbl}
\usepackage[nohyperlinks]{acronym}
\usepackage[abbreviations]{foreign} 
\usepackage{pifont}
\usepackage{adjustbox}

\usepackage{array}
\usepackage{subcaption}
\definecolor{Gray}{gray}{0.8}
\definecolor{LightCyan}{rgb}{0.88,1,1}

\title{Survey of Federated Learning Models for Spatial-Temporal Mobility Applications}

\author{ \href{https://orcid.org/0000-0003-3433-1337}{\includegraphics[scale=0.06]{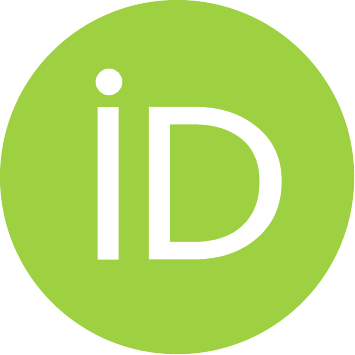}\hspace{1mm}Yacine Belal}\\
	National Institute of Applied Sciences\\
	Lyon, France \\
	\texttt{yacine.belal@insa-lyon.fr} \\
 	\And
	\href{https://orcid.org/0000-0003-2821-7714}{\includegraphics[scale=0.06]{orcid.pdf}\hspace{1mm}Sonia Ben Mokhtar} \\
	National Institute of Applied Sciences\\
	CNRS\\
        Lyon, France \\
	\texttt{sonia.benmokhtar@insa-lyon.fr} \\
        \And 
        \href{https://orcid.org/0000-0002-5895-8903}{\includegraphics[scale=0.06]{orcid.pdf}\hspace{1mm}Hamed Haddadi} \\
	Imperial College London\\
        London, UK \\
	\texttt{h.haddadi@imperial.ac.uk} \\  
        \And
        Jaron Wang \\
	University of Washington\\
        Bothell, USA \\
	\texttt{jierui98@uw.edu} \\
        \And
        \href{https://orcid.org/0000-0003-4631-4438}{\includegraphics[scale=0.06]{orcid.pdf}\hspace{1mm}Afra Mashhadi} \\
	University of Washington\\
        Bothell, USA \\
	\texttt{mashhadi@uw.edu} \\}

\hypersetup{
pdftitle={SURVEY OF FEDERATED LEARNING MODELS FOR
SPATIAL-TEMPORAL MOBILITY APPLICATIONS},
pdfsubject={cs.LG},
pdfauthor={Yacine Belal, Sonia Ben Mokhtar, Hamed Haddadi, Afra Mashhadi},
pdfkeywords={Federated Learning, Spatial-Temporal Mobility, Privacy-preserving Learning, Flow prediction, Point-of-Interest recommendation},
}

\begin{document}
\maketitle

\begin{abstract}
Federated learning involves training statistical models over edge devices such as mobile phones such that the training data is kept local. Federated Learning (FL) can serve as an ideal candidate for training spatial temporal models that rely on  heterogeneous and potentially massive numbers of participants while preserving the privacy of highly sensitive location data. However, there are unique challenges involved with transitioning existing spatial temporal models to decentralized learning. In this survey paper, we review the existing literature that has proposed FL-based models for predicting human mobility, traffic prediction, community detection,  location-based recommendation systems, and other spatial-temporal tasks. We describe the metrics and datasets these works have been using and create a baseline  of these approaches in comparison to the centralized settings. Finally, we discuss the challenges of applying spatial-temporal models in a decentralized setting and by highlighting the gaps in the literature we provide a road map and opportunities for the research community.
\end{abstract}


\section{Introduction}
Spatio-temporal mobility data collected by location-based services (LBS)~\cite{huang2018location} and other means such as Call Data Records (CDR), WiFi hotspots, smart watches, cars, etc. is very useful  from a socio-economical perspective as it is at the heart of many useful applications (e.g., navigation, geo-located search, geo-located games) and it allows answering numerous societal research questions~\cite{kolodziej2017local}. For example, CDR has been successfully used to provide real-time traffic anomaly as well as event detection \cite{toch2019analyzing, wang2020deep}, and a variety of mobility datasets have been used in shaping policies for urban communities~\cite{ferreira_deep_2020} or epidemic management in the public health domain~\cite{oliver2015mobile,oliver2020mobile}. From an individual-level perspective, users can benefit from personalized recommendations when they are encouraged to share their location data with third parties~\cite{erdemir_privacy-aware_2020}. 

While there is no doubt about the usefulness of location-based applications, privacy concerns regarding the collection and sharing of individuals’ mobility traces or aggregated flow of movements have prevented the data from being utilized to their full potential~\cite{shokri2011quantifying, beresford2003location, krumm2009survey}. A mobility privacy study conducted by De Montjoye et al. \cite{de2013unique} illustrates that four spatio-temporal points are enough to identify 95\% of the individuals, which exacerbates the user re-identification risk and could be the origin of many unexpected privacy leakages. Indeed, various studies have shown that numerous threats arise   if location data falls into the hands of inappropriate parties. These threats include re-identification~\cite{maouche2017ap}, the inference of sensitive information about users \cite{krumm2009survey,wernke2014classification}(e.g., their home and work locations, religious beliefs, political interests or sexual orientation). In some extreme cases, sharing geo-located data may even endanger users' physical integrity (e.g., the identification of protesters in dictatorial regimes or during wars\footnote{https://www.independent.co.uk/news/ukraine-ap-russia-gps-kyiv-b2093310.html}) or their belongings (e.g., robbery\footnote{https://pleaserobme.com}). 

One way to consider  addressing privacy challenges is to break from centralized data collection and maintain the location data on user devices. In  this decentralized paradigm,  a variety of solutions are plausible.  A possible approach is to  create accurate enough synthetic location data to  train spatial-temporal models and then fine-tune them on users' devices to perform best for the   individuals' location data.  Although this is a promising approach, the research has shown that mobility trajectory generation is an extremely difficult task that may not portray data heterogeneity accurately~\cite{kong2023mobility,long2023practical}. Moreover,  trajectories from different locations,  periods, or user groups may have distinct characteristics that the pre-trained model does not handle effectively. Although this is an ongoing field of research, to date there exists little evidence as to whether models can transfer mobility knowledge across cities~\cite{he2020human}, as with a significant distribution shift between the data used for pre-training and the data for the target task, transfer learning may not generalize well.

In parallel, an alternative approach that is increasingly considered a promising approach is  Federated Learning (FL)~\cite{mcmahan2016federated}. FL relies on clients (be it users' smartphones or edge devices) to train a machine learning model on their local training data and share the model weights with a central server (called the FL server) that aggregates the received clients' contributions. As such, FL empowers clients by allowing them to benefit from a globally trained model while keeping their private data on their premises. Compared with other applications of FL, spatial-temporal data and models possess certain characteristics and properties  that introduce challenges. For instance, in domains involving ST data, the observations made at nearby times and locations  cannot hold the independent property due to the auto-correlation (e.g., changes in traffic activity occur smoothly over time and space). As a result, classical  FL algorithms that assume independence among observations are not suited for ST applications~\cite{atluri_spatio-temporal_2018}. Furthermore, ST data is not identically distributed, that is it does not meet the assumption that    every instance belongs to the same population and is thus identically distributed. These two properties are referred to as \textit{iid} (independent and identically distributed). Additionally, ST data is often  both spatially and or temporally sparse and often created  at high frequencies, especially in real-time applications, resulting in a high velocity and volume of data generation.

\begin{figure}
    \centering  \includegraphics[width=0.56\linewidth]{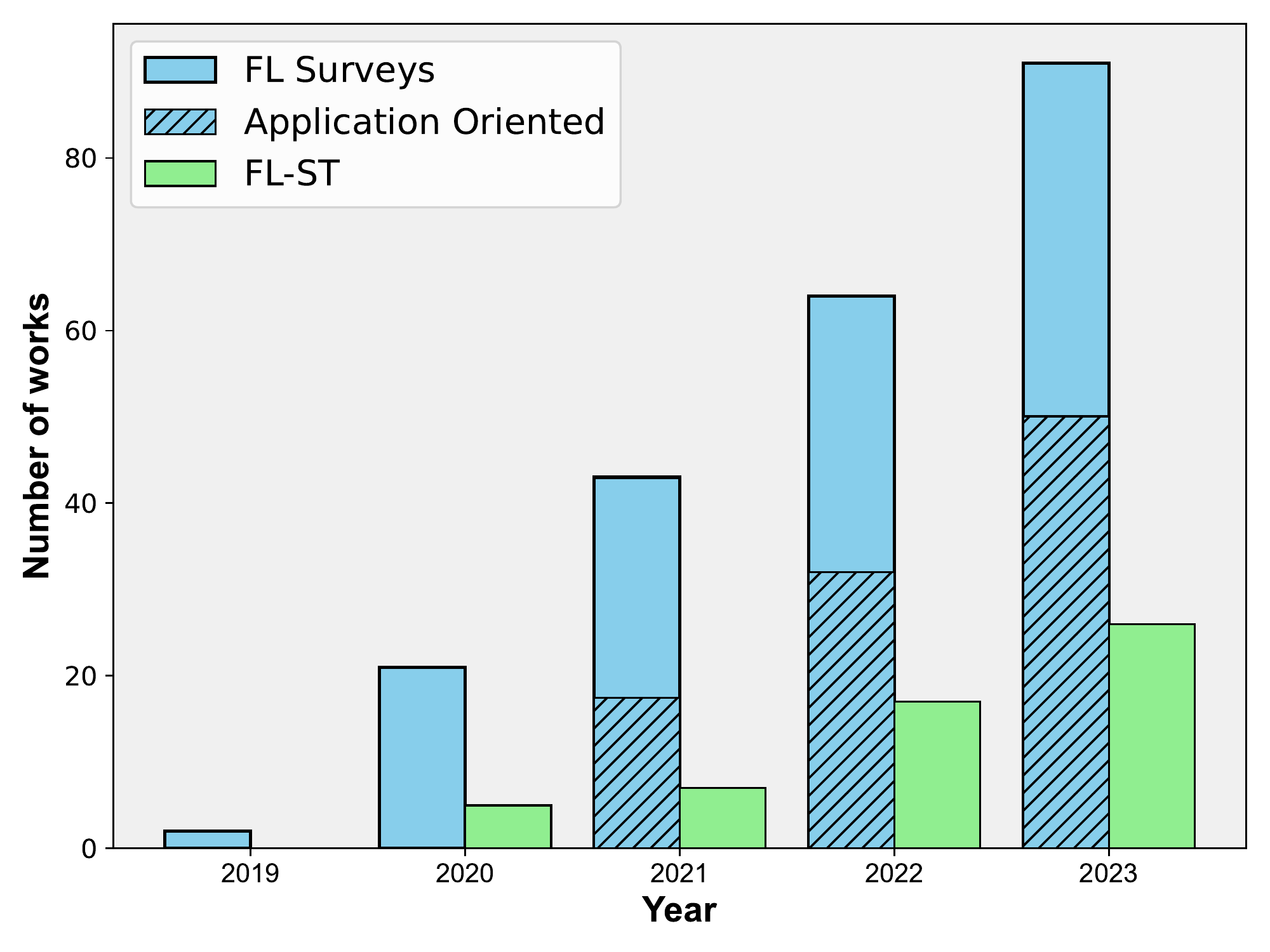}
    \caption{Number of FL Surveys in comparison with the growing number of spatial-temporal FL methodological papers.  The shaded areas present the percentage of the surveys that are application specific.}
    \label{fig:survey}
\end{figure}

\paragraph{\textbf{Related Surveys}}
Since FL emerged half a decade ago,  many survey papers~\cite{li2021survey,wen2023survey,du2020federated,fu2022federated,mothukuri2021survey,rasha2023federated} have reviewed relevant literature from various perspectives.  In 2023 alone, 91 surveys on federated learning were published with  the majority being on applications-specific topics, predominantly focused on  health, finance, blockchain, and the Internet of Things (IoT). Figure~\ref{fig:survey} presents the growing number of application-specific FL surveys and in parallel the emergence of papers on FL-ST topics\footnote{Our methodology in harvesting data about all survey papers on FL is outlined in the Appendix.}. 
In motivating the need for this survey, we discuss  the perspectives that IoT-related and generic FL surveys have covered and highlight their shortcomings  when applied to ST models.  More specifically, FL Surveys on smart cities including those of IoT applications are closest to our work.  Zheng et al.~\cite{zheng2022applications} reviewed applications of FL in smart cities. They provided a broad overview of various applications of smart cities including IoT, transportation, communication, medical care, and finance. They created a glossary of the papers in these domains without discussing the diversity of existing approaches or drawing comparisons between these works. Their work mainly focused on security and privacy challenges and outlined algorithmic efficiency as a future direction but did not consider the heterogeneity and sparsity  often associated with the location data. The closest survey to ours is~\cite{pandya2023federated} which has a subsection  discussing the FL transportation application in smart cities. However, they focus on car navigation, number plate recognition, railway, and infrastructure systems and reviewed 6 papers in this domain out of which only one falls in the focus of our survey. 

Other FL surveys that are \textit{not} application-specific and review methodological approaches of FL regarding heterogeneity, model convergence, and personalization fall short in drawing a comparison between these different  techniques when applied to ST data. Although some of the challenges such as non-iid are well covered in those surveys,  the other characteristics of the ST data are not covered. Finally, we note that despite its unique properties  and the increasing number of spatial-temporal research papers emerging in  FL (Figure~\ref{fig:survey}), no survey paper has attempted to categorize, review, and compare this emergent of ST FL methodological approaches.

\paragraph{Our Contribution}
The contributions of this survey are as follows:
\begin{itemize}
    \item In this survey, we study 38 existing spatial-temporal works that have been adapted to the FL paradigm and discuss their strengths and limitations. In reviewing these recent works, we give a concise but concrete description of each approach along with a comparison of the baseline datasets. 

    \item As part of this survey we create an open-source repository\footnote{https://github.com/YacineBelal/Survey-of-Federated-Learning-Models-for-Spatial-Temporal-Mobility-Applications} where we host the existent FL-ST algorithms that are available as well as sample datasets. We also include available mobility data handling (preprocessing, analysis, and visualization) software. We hope that this open-source repository enables the research community to contribute their work and enable a greater number of baselines. 
    \item  We discuss  open challenges and a roadmap that we envision for the research community to explore in the coming years. Some of the research questions that emerge from our survey are: \textit{How can the ST mobility research community establish and promote standardized practices? How can FL effectively achieve better personalization through the integration of semantic information?~(See \cref{subsec:semantic}) What novel techniques and solutions can be developed to detect and mitigate Byzantine behaviors in ST mobility FL? How can current ST mobility frameworks be optimized to enhance communication efficiency?~(See \cref{subsec:communication})}
\end{itemize}

\paragraph{Organization of this survey}

 The remainder of this paper is structured as follows. First, we present a background on FL in Section~\ref{sec:fl}. Then, we present a set of spatial-temporal applications under study in Section~\ref{sec:applications} and detail the used FL approaches for these applications in Section~\ref{sec:approaches}. Finally, we discuss open research challenges in Section~\ref{sec:futur} and conclude the paper in Section~\ref{sec:conclusion}.

\section{Federated Learning}\label{sec:fl}

\begin{figure}
    \centering
    \includegraphics[width=\textwidth]{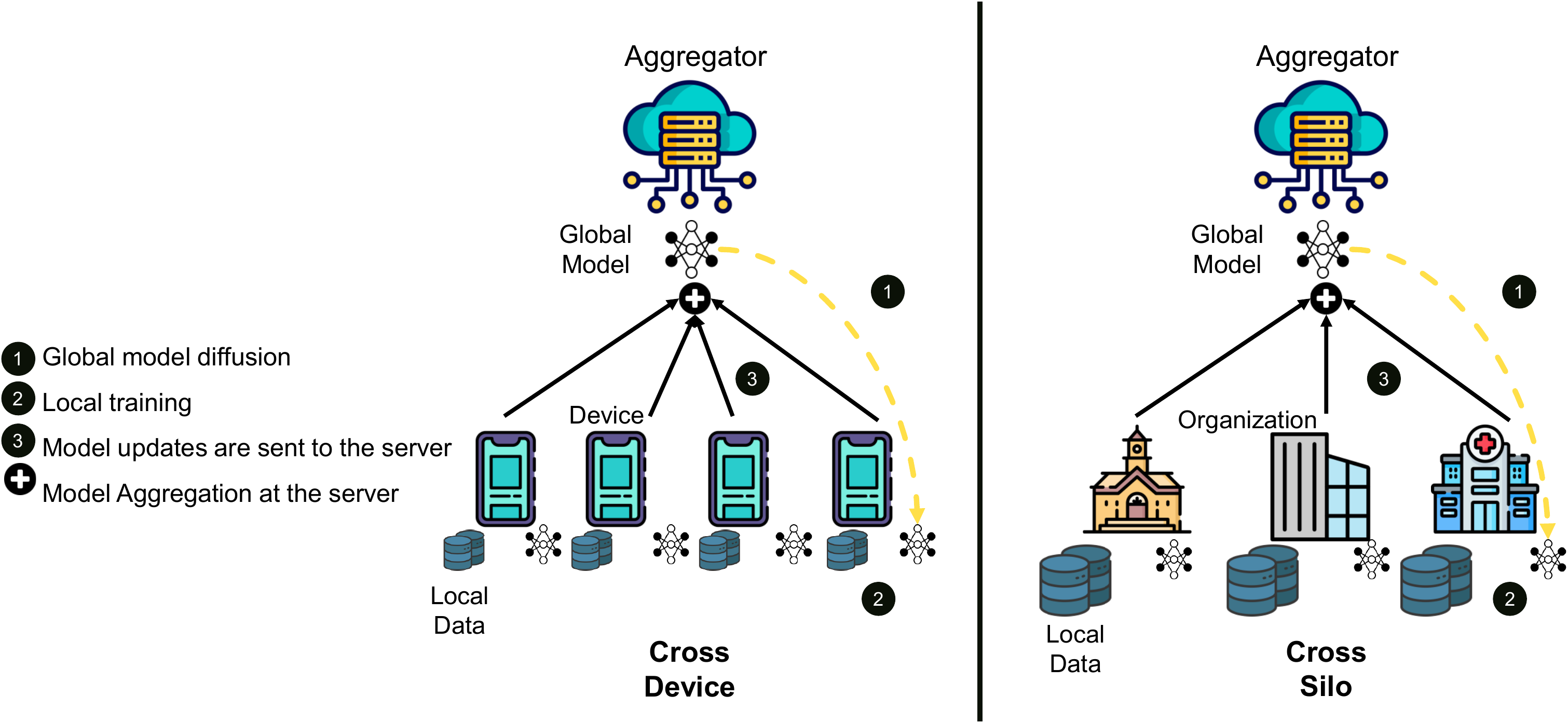}
    \caption{Federated Learning general workflow. On the left, a cross-device setting where each client is a device (\eg, smartphone) and on the right is cross-silo setting where each client represents an organization. The main difference between the two is the magnitude both in terms of clients and local data.}
    \label{fig:fl}
\end{figure}

\emph{Federated Learning (FL)} is a paradigm to perform distributed machine learning at the edge~\cite{47976,kairouz2019advances,fedlearn_1, fedlearn_2}. 
In FL, a global (joint) model is trained in a decentralized fashion without the need to collect and process user data centrally. The central service provider, often called aggregator, distributes a shared ML model to multiple users for training on local data, and then aggregates the resulting models into a single, more powerful model, using an aggregation method~(\eg, Federated Averaging~\cite{fedlearn_1}). Figure~\ref{fig:fl} illustrates an example of FL architecture. 

More formally, let $\theta$ be the global model that an FL instance aims to learn and $F(\theta)$ be the cost function that evaluates $\theta$. let $\mathcal{N} = \{1, 2, \ldots, n\}$ be the set of users, with each user $i$ possessing a local dataset $\mathcal{D}_i$. Note that we use 'users' and 'participants' interchangeably throughout the paper. 

The aggregator initiates by randomly sampling $\mathcal{S}= \{1, 2, \ldots, m\}$ with $\mathcal{S} \subset \mathcal{N}$ being a subset of users chosen to participate to a specific round $t$. The aggregator transmits the global model $\theta^t$ to $\mathcal{S}$. 
Each selected participant $i$ trains $\theta^t$ on $\mathcal{D}_i$ to obtain an updated model, denoted $\theta_i^{t+1}$. Then, $i$ send the model updates $\nabla F(\theta_t,\mathcal{D}_i)$ (or $\theta_i^{t+1}$) back to the aggregator (see notation in Table~\ref{tab:fl_definitions}). Communication overhead can be reduced by applying a random mask to the model weights~\cite{fedlearn_2}. The central server then aggregates the received updates~\cite{reddi2020adaptive} to create the new global model, $\theta^{t+1}$. This round-based mechanism is repeated until some convergence criteria of $\theta$ is reached. Equation~\ref{eq:1} presents a Federated Averaging-based aggregation step (See Table~\ref{tab:fl_definitions} in Appendix for notation).

\begin{equation}
\label{eq:1}
\theta^{t+1} = \theta^{t} - \frac{\eta}{\sum_{i=1}^{m} |\mathcal{D}_i|}\sum_{i=1}^m |\mathcal{D}_i| \cdot \nabla F(\theta_t,\mathcal{D}_i)
\end{equation}

Where $\eta$ denotes the global learning rate and $|.|$ denotes the set cardinality operation.



This described algorithm represents the general FL workflow. However, it is essential to note that there exist several flavors of FL. Based on the order of magnitude of the number of participants and the assumptions regarding their computational capabilities, two FL settings are distinguished in this paper: cross-device FL~(CD FL) and cross-silo FL~(CS FL). 

CD FL typically refers to scenarios where the number of users is in the order of millions of individuals, each possessing limited and heterogeneous computational resources (\eg, personal laptops, cell phones). In contrast, cross-silo FL operates at a (much) smaller scale of the number of users, representing each a large entity (\eg, hospitals, cellular operators, and data centers), with substantial computational capabilities. Figure~\ref{fig:fl} presents an overview of both schemes of FL.


\subsection{Challenges}
There are various challenges in FL that the research community has been studying in the past years (see~\cite{kairouz2021advances} for full reference).  Two of these challenges impacting FL models are generally regarding \textbf{data heterogeneity} and \textbf{model personalization}.

\begin{itemize}
\item  Data heterogeneity refers to the scenario where clients' data are not identically and independently distributed (IID) which could lead to domain shift problems making learning a generalizable representation a difficult task. Consider two users $i$ and $j$, with data following probability distributions $P_i$ and $P_j$, respectively. These probabilities are typically rewritten as
$P_i(Y|X)P_i(X)$ and $P_j(Y|X)P_j(X)$, respectively, where $X$ and $Y$ are the features' (\eg, visited POIs) and labels' (\eg, next POI) probability distributions. Considering this notation, data can be non-IID in several ways: the marginal distribution of features can vary between clients (\ie, $P_i(X) \neq P_j(X)$), such is the case for clients who have significantly different preferences in a POI recommendation use case. This is known as a covariate shift. Alternatively, the marginal distribution of labels can be different, also known as the prior distribution shift
(\ie, $P_i(Y) \neq P_j(Y)$). This is especially the case for clients tied to different geo-regions. Finally, there can also be a concept shift, that is, the conditional distributions $P(X|Y)$ and $P(Y|X)$ can highly vary across clients and context (\eg, weather, time... etc). As we will discuss in the next sections, the human trajectory is extremely unique~\cite{de2013unique} and strongly influenced by different meta-parameters. This leads to the coexistence of different types of non-iid property and calls for an increased need for personalization. 
\item Model Personalization on the other hand refers to the process of assigning different model parameters (weights) to different clients. The objective behind this is twofold: i)  improve the global model's individual performance and ii) mitigate the impact of non-iid property. To this end, various methodologies have emerged. One approach involves incorporating context features into the model during training, known as personalization through featurization. Another technique, meta-learning, focuses on adapting the global model to the local learning task. In its simplest form, it boils down to fine-tuning the global model or a subset of layers of it. Understanding what level of generalization can be learned globally and what layers of the models need to be personalized locally is an active area of research.
\end{itemize}

There are several other FL challenges that we consider throughout this paper. Firstly, concerns over data and model parameter privacy, which are addressed through methods such as differential privacy~\cite{fedlearn_dp}, Trusted Execution Environments~\cite{10.1145/3458864.3466628}, or encryption~\cite{bonawitz2017practical, hardy2017private} (i). Secondly, Byzantine Resilience, defined as the capability to train an accurate statistical model despite arbitrary behaviors, presents another significant challenge (ii). Finally, Communication Efficiency stands as a crucial aspect to consider (iii).

In addition to these challenges, there are specific considerations within the realm of spatio-temporal mobility applications. These include: 1) the constraints on client resources, 2) the online learning setting - specifically, whether clients train while simultaneously collecting/creating data, and 3) the comparison of solutions with at least one federated competitor, aside from standalone FL. This comprehensive evaluation framework has allowed us to conduct a meta-comparison, analyzing the strengths and limitations of the main works within each approach~(Refer to~\cref{tab:npr,tab:tfp-grid,tab:tfp-gnn,tab:recsys}). Moreover, it has facilitated the identification of potential research directions~(See \cref{sec:futur}).


\subsection{Frameworks}

As Federated Learning (FL) establishes itself as the standard machine learning paradigm, numerous frameworks have emerged to offer scalable and flexible solutions for privacy-preserving machine learning in distributed settings. Selecting an appropriate FL framework for designing new algorithms is a non-trivial task. Therefore, in this section, we define specific criteria to facilitate the selection process of an FL framework. The following criteria will be considered:

\begin{itemize}
    \item \textbf{Supported Frameworks}: 
    Most FL frameworks can be extended to integrate with various machine learning frameworks. However, this process is often complex. Hence, we will outline the machine learning frameworks inherently supported by each FL framework and we will qualify by "native" each framework that provides its own optimization algorithms. 
    \item \textbf{Privacy tools}: 
    As privacy stands as a central motivation for FL, it is imperative to recognize that standalone FL lacks privacy~\cite{rasha2023federated,kairouz2021advances}. Within this context, various privacy technologies are under design, including cryptographic tools like Secure Multi-Party Computation (SMPC), Homomorphic Encryption (HE), secret sharing, formal models such as Differential Privacy (DP), and hardware-based techniques like Trusted-Execution Environments (TEEs). Evaluating which of these tools an FL framework supports is crucial due to the complexity of integrating them.
    \item \textbf{Scenario}:
    This criterion assesses the framework's capability to support real-world deployments or whether it is exclusively designed for simulations.
    \item \textbf{Benchmarking}: 
    To make the research-work pipeline more efficient, most FL frameworks are packaged with pre-defined machine learning tasks, encompassing models and datasets, along with benchmarks. These tasks will be highlighted for each framework. It's important to note that while frameworks come with suggested tasks, they are not restricted solely to those tasks.
\end{itemize}
Table~\ref{tab:frameworks} presents a summary of the specified evaluation criteria for the main FL frameworks. Notably, most frameworks are extensively benchmarked for tasks in computer vision and Natural Language Processing (NLP), with relatively limited consideration for spatial mobility applications. \textit{FedScale~\cite{lai2022fedscale} distinguishes itself by offering starting pack mobility datasets, }including a task for taxi trajectory prediction utilizing the TaxiPorto dataset~\cite{pkdd-15-predict-taxi-service-trajectory-i}. Additionally, it incorporates the Waymo Motion Dataset, comprising object trajectories and corresponding 3D maps for 103,354 scenes, utilized in autonomous driving scenarios.

\textit{A noteworthy feature is observed in FederatedScope~\cite{xie2022federatedscope}, which introduces benchmarks for graph learning}, potentially applicable to Graph Neural Network (GNN)-based approaches for flow prediction tasks. 

Regarding machine learning framework support, it's notable that only two of the existing solutions offer a framework-agnostic approach to Federated Learning—namely, Flower~\cite{beutel2020flower} and FedML~\cite{he2020fedml}. 

Considering privacy enhancements, most frameworks integrate DP and SMPC functionalities. However, LEAF~\cite{caldas2018leaf} does not provide support for privacy-enhancing technologies. On the other hand, FATE~\cite{liu2021fate} stands out by offering unique Homomorphic Encryption support.

For an in-depth comparative analysis across the described FL frameworks, a comprehensive comparison is available in~\cite{liu2022unifed}, while extensive comparisons of peer-to-peer FL frameworks have been conducted in~\cite{beltran2022decentralized}.

\begin{table}[!h]
\centering\captionsetup{justification = centering}
\caption{Available functionalities and bench-marking across different FL frameworks.}
\label{tab:frameworks}
\begin{tabular}{|c|c|c|c|c|}
\rowcolor{Gray}
\hline 
Name &
  Supported Frameworks &
  \textbf{Privacy tools} &
  Scenario &
 \textbf{Benchmarking}\\ \hline 
Tensorflow Federated (TFF) &
  Tensorflow &
  \begin{tabular}[c]{@{}c@{}}DP\\ SMPC\end{tabular} &
  Simulation &
  \begin{tabular}[c]{@{}c@{}}Computer Vision\\ NLP\end{tabular} \\ \hline
Pysift \& Pygrid &
  \begin{tabular}[c]{@{}c@{}}Tensorflow\\ Pytorch\end{tabular} &
  \begin{tabular}[c]{@{}c@{}}DP\\ SMPC\end{tabular} &
  \begin{tabular}[c]{@{}c@{}}Simulation\\ Real\end{tabular} &
  Computer Vision \\ \hline
FATE &
  \begin{tabular}[c]{@{}c@{}}Tensorflow\\ Pytorch\\ Sickit-learn\end{tabular} &
  HE &
  \begin{tabular}[c]{@{}c@{}}Simulation\\ Real\end{tabular} &
  Computer Vision \\ \hline
Flower &
  Framework-agnostic &
  DP &
  Simulation &
  Computer Vision \\ \hline
PaddleFL &
  \begin{tabular}[c]{@{}c@{}}Native\end{tabular} &
  \begin{tabular}[c]{@{}c@{}}DP\\ SMPC\\ Secret Sharing\end{tabular} &
  \begin{tabular}[c]{@{}c@{}}Simulation\\ Real\end{tabular} &
  \begin{tabular}[c]{@{}c@{}}Computer Vision \\ NLP\\ Recommendation\end{tabular} \\ \hline
FederatedScope &
  \begin{tabular}[c]{@{}c@{}}Tensorflow\\ Pytorch\end{tabular} &
  \begin{tabular}[c]{@{}c@{}}DP\\ SMPC\end{tabular} &
  \begin{tabular}[c]{@{}c@{}}Simulation\\ Real\end{tabular} &
  \begin{tabular}[c]{@{}c@{}}Computer Vision\\ NLP\\ Recommendation\\ \textbf{Graph Learning}\end{tabular} \\ \hline
LEAF &
  \begin{tabular}[c]{@{}c@{}}Tensorflow\end{tabular} &
  / &
  Simulation &
  \begin{tabular}[c]{@{}c@{}}Computer Vision\\ NLP\end{tabular} \\ \hline
FedML &
  Framework-agnostic&
  SMPC &
  \begin{tabular}[c]{@{}c@{}}Simulation\\ Real\end{tabular} &
  \begin{tabular}[c]{@{}c@{}}Computer Vision\\ NLP\end{tabular} \\ \hline
FedScale &
  \begin{tabular}[c]{@{}c@{}}Tensorflow\\ Pytorch\end{tabular} &
  \begin{tabular}[c]{@{}c@{}}DP\\ SMPC\end{tabular} &
  \begin{tabular}[c]{@{}c@{}}Simulation\\ Real\end{tabular} &
  \begin{tabular}[c]{@{}c@{}}Computer Vision\\ NLP\\ Recommendation\\ \textbf{Mobility Prediction}\\ RL\end{tabular} \\ \hline
\end{tabular}
\end{table}

\section{Applications}\label{sec:applications}

\paragraph{\textbf{Mobility Prediction}}

Mobility prediction can be defined as algorithms and techniques  to estimate the future locations of users. Predicting the next location of users can help with a range of applications including networking (e.g.,  handover management),  pandemic management (e.g., contact tracing), etc.   This type of prediction is performed on individual users' traces where the historical trend of the user's visited locations can help in predicting the likelihood of their next location.

\paragraph{\textbf{Transportation}}
With the rising availability of transportation data collected from various sensors like   road cameras, GPS probes, and IoT devices,  there is an enormous opportunity for city planners to leverage these types of data to  facilitate various tasks such as traffic flow prediction.  Different from \textit{trajectory} data that records a sequence of locations and time in each trip, crowd flow data only have the start and end locations of a trip, and how many people flow in and out of a particular region can be counted. Indeed, traffic flow prediction using spatial-temporal data has been  one of the main focuses of the research community (See comprehensive survey~\cite{jiang2021dl}). In the context of transportation, this problem  is often considered as forecasting which is to predict traffic speed or traffic flow of regions or road segments based on historical aggregated mobility data. 


\paragraph{\textbf{Community Detection}}
Community detection  is an  important aspect of urban planning as it allows researchers and planners to identify patterns and trends in human movement. By identifying groups of individuals or locations that are highly connected, researchers and planners can gain insight into how people move through a city, which can inform the design of transportation systems and urban spaces. Additionally, community detection can help identify areas of a city that are at risk of overcrowding or under-utilization, allowing for proactive measures to be taken to address these issues.  The underlying enabler of identifying  urban communities~\cite{ghahramani2018extracting,ferreira_deep_2020} is spatial temporal data that presents  the amount of time spent in different parts of the city. Researchers have  shown that human mobility exhibits a strong  degree of non-linearity~\cite{de2013interdependence} and models that rely on non-linear clustering algorithms, such as the one proposed by Ferreira et al.~\cite{ferreira2020deep},  to detect urban communities have been shown to outperform traditional approaches such as  Principal Component Analysis (PCA), and Model-Based Clustering (MBC) and DB-SCAN techniques on a variety of centralized geospatial traces. 

\paragraph{\textbf{Location Based Social Networks}}

LBSNs such as Foursquare and Flickr are social networks that use GPS features to locate the users and let the users broadcast their locations and other content from their mobile devices. LBSNs do not merely mean that the locations are added to the user-generated content (UGC) in social networks so that people can share their location information but also reshape the social structure among individual users that are connected by both their locations in the physical world and their location-tagged social media content in the virtual world. LBSNs contain a large number of user check-in data which consists of the instant locations of each user. Such social networks could also be thought of as the underlying application of Location-based recommendation systems.

\section{Approaches}\label{sec:approaches}

In this section we summarize federated learning spatial-temporal approaches in three main categories of i)  trajectory predictive approaches which focus on the next-point prediction of user's trajectories, ii) traffic flow prediction approaches, and iii) clustering approaches. For each approach, we present the evaluation metrics, the datasets used and the various FL strategies.  Table~\ref{tab:stoa} provides an overview of the most important works. 

\begin{table*}[!ht]
\centering
\caption{CD denotes Cross-Device approaches and CS for cross-silo}
\label{tab:stoa}
\begin{tabular}{|c|c|p{3cm}|p{4cm}|c|}
\rowcolor{Gray}
\hline
\textbf{Model}  & Year &  Approach     & \textbf{Dataset} & \textbf{Federated Strategy } \\
\hline
\multicolumn{5}{|c|}{\textbf{\textit{Trajectory Predictive Approaches}}}\\
\hline
    \citeauthor{fan2019decentralized}~\cite{fan2019decentralized} &2019   &  Transfer-learning & Private Mobile Phone Traces  & CD \\
        \hline
   
     PMF~\cite{feng2020pmf} & 2020 & &Foursquare~\cite{yang2014modeling},~ DenseGPS~\cite{feng2018deepmove}, Twitter~\cite{zhang2017regions} & CD, Attack-resilient \\
     \hline
     

      STSAN~\cite{li2020predicting} & 2020 & ST Attention Layer &    Foursquare\cite{yang2014modeling}, Twitter~\cite{zhang2016gmove} & CD
    \\
    & &    & Yelp~\cite{liu2017experimental} & Adaptive Model Fusion  \\
     \hline
~\citeauthor{ezequiel2022federated}\cite{ezequiel2022federated} &2022   & GRU-Spatial and Flashback  & Foursquare~\cite{yang2014modeling}, Gowalla~\cite{gowalla} &  CD \\
    \hline

     STLPF~\cite{wang2022location} & 2022 &   AutoEncoder with    & Foursquare\cite{yang2014modeling} & CD\\
    & & Global/Local attention&  & \\
     \hline
\multicolumn{5}{|c|}{\textbf{\textit{Flow Predictive Approaches }}}\\  
     \hline
     FedGRU~\cite{liu_privacy_2020} & 2020 & GRU &  PeMS~\cite{chen2001freeway} & CS, FedAvg \\
     
        \hline
        FedTSE~\cite{yuan2022fedtse} & 2022 & Reinforcement Learning & England Freeway Dataset\cite{england2017traffic} &  CS, FedAvg  \\
        
        \hline
     FedSTN~\cite{yuan2022fedstn}  & 2022 &  GNN   &  Taxi-NYC~\cite{taxinyc2009}, Taxi-BJ~\cite{jagadish2014big} & CS, Vertical FL\\
     \hline
     CNFGNN  ~\cite{meng2021cross}  & 2021 & GNN  & PeMS-BAY~\cite{li2017diffusion}, METR-LA~\cite{jagadish2014big} & CS\\
     CTFL~\cite{zhang2022graph} & 2022 & GNN & PeMSD4, PeMSD7 & Clustered FL,  CS\\
     MVFF~\cite{errounda2022mobility} & 2022 & GRU+GNN & Yelp~\cite{liu2017experimental}, NY-Bike~\cite{nycbike2013} & Vertical FL, CS\\
\hline
\multicolumn{5}{|c|}{\textbf{\textit{Community Detection Approaches}}}\\
\hline
     F-DEC~\cite{mashhadi2021deep}  & 2021 & Deep Embedded Clustering    & GeoLife~\cite{zheng2010geolife} & CD, FedAvg  \\
     \hline
\multicolumn{5}{|c|}{\textbf{Other}}\\  
\hline
EDEN~\cite{khalfoun2021eden}  & 2021 & \multicolumn{2}{|c|}{Privacy Optimization} &  CD\\
PREFER~\cite{guo2021prefer}  & 2021 & \multicolumn{2}{|c|}{Location Rec Sys} & CD\\
PEPPER~\cite{belal2022pepper}  & 2022 & \multicolumn{2}{|c|}{Location Rec Sys} & CD, Gossip Learning\\
MTSSFL~\cite{zhang2021toward} & 2021 &\multicolumn{2}{|c|}{ Transport Mode Inference} &  CD\\
 Fed-DA~\cite{zhang2021dual} & 2021 & \multicolumn{2}{|c|}{Network Traffic} & CS \\
 Fed-NTP~\cite{sepasgozar2022fed}  & 2022 & \multicolumn{2}{|c|}{Network Traffic}  &  CS\\

\hline
\end{tabular}
\end{table*}

\subsection{Trajectory Predictive Approaches}

Given a user’s trajectory, these approaches  aim at predicting the user’s next position.   In a centralized setting where the training data containing all users' trajectories are available, RNN-based approaches including LSTM and GRU can be broadly applied in dealing with trajectories and predictive tasks~\cite{liao2018predicting,feng2018deepmove,gao2019predicting,liu2016predicting}.  
As this type of data (i.e., trajectory) is highly privacy sensitive, one of the challenges that the research community has been focusing on  creating models that can be tuned for privacy and utility, namely, PUTs (Privacy-Utility Tradeoff). Models such as \cite{zhan2022privacy,rao2020lstm,erdemir_privacy-aware_2020} leverage the centralized training data to enhance the privacy level of the traces by reducing user's re-identification and at the same time optimizing for the utility of the predictions (i.e., higher accuracy of next point predictions). 

The challenges of trajectory predictive approaches in decentralized learning are different. Firstly, the unique properties of people's mobility~\cite{de2013unique}  lead to a non-iid distribution of the data amongst clients. Second and as a result of the first challenge, creating a global  model for predicting the next location of users that works equally well for all the users becomes an extremely challenging task. That is  one must decide to what extent should clients adopt the global model and when to opt-in for a purely personalized model.   In this section we review the existing works in this domain, review how they account for the mentioned challenge, and compare their performances in Table~\ref{tab:acc} against    centralized predictive approaches namely ST-RNN~\cite{liu2016predicting}, MCARNN~\cite{liao2018predicting}, DeepMove~\cite{feng2018deepmove}, and VANext~\cite{gao2019predicting}.
 
\begin{table}[ht]
\centering
\caption{Baselines for next location prediction models}
\label{tab:acc}
\begin{tabular}{lll|llll}
         & \multicolumn{2}{c}{Foursquare NY} & \multicolumn{2}{c}{Foursquare Tokyo}   \\
         \hline
         & Acc@1 (ACC@5)   & APR    & Acc@1 (ACC@5)   & APR        \\
         \hline
         \multicolumn{5}{c}{Centralized Baselines}\\
         \hline
ST-RNN~\cite{liu2016predicting}   & 0.2633 & 0.9431 & 0.2567 & 0.9536 &    \\
MCARNN~\cite{liao2018predicting}  & 0.3167 & 0.9595 & 0.2770 & 0.9532 &    \\
DeepMOVE~\cite{feng2018deepmove} & 0.3010 & 0.9221 & 0.2668 & 0.9257 &    \\
VaNext~\cite{gao2019predicting}   & 0.3627 & 0.9792 & 0.3436 & 0.9735 &    \\
\hline
       \multicolumn{5}{c}{Federated Approaches}\\
       \hline
STSAN~\cite{li2020federated}    & 0.4297 & 0.9902 & 0.3906 & 0.9847 &    \\
STLPF~\cite{wang2022location}    & 0.4067 & 0.9893 & 0.3887 & 0.9856 &   \\
PMF~\cite{feng2020pmf}      &   NA     &    NA    & 0.2130  & NA     &       \\
Ezequiel~\cite{ezequiel2022federated} & 0.1133  &   NA     &      NA  &    NA    &   \\      
\end{tabular}
\end{table}


\subsubsection{Federated Trajectory Predictive Approaches}
\Cref{tab:npr} offers a concise overview of the challenges addressed by the existing literature in trajectory predictive approaches. In this application domain, the non-iid nature stemming from a substantial number of users, each with unique characteristics, as well as privacy considerations, takes precedence over other challenges. Notably, while the former challenge appears to have been adequately addressed by the majority of the reviewed works, the latter has only been specifically tackled by\cite{feng2020pmf}. In addition, it is noteworthy that most works have not taken into account federated competitors, preventing a comprehensive evaluation of their impact. This concern is pivotal not only for the current task but for all spatio-temporal mobility FL, as elaborated in ~\Cref{subsec:standardization}. Similar apprehensions extend to the robustness against Byzantine behaviors. However, this criterion holds less significance in the context of CD FL compared to the CS FL setting, as is the case for the next discussed approach. In the following, we delve into a more comprehensive discussion of these works.

\begin{table}[!ht]
\caption{Summary of the challenges tackled by the reviewed Trajectory Predictive works.}\label{tab:npr}
\begin{tabular}{rrrrrrrr}
\hline
                                                    & Privacy                    & \begin{tabular}[c]{@{}c@{}}Byzantine\\ Resilience\end{tabular} & Non-IIDness                & \begin{tabular}[c]{@{}c@{}}Resource\\ constraints\end{tabular} & \begin{tabular}[c]{@{}c@{}}Overhead \\ assessment\end{tabular} & \begin{tabular}[c]{@{}c@{}}Federated\\ competitors\end{tabular} & \begin{tabular}[c]{@{}c@{}}Online\\ Learning\end{tabular} \\ \hline
\citet{fan2019decentralized}       & \ding{55} & \ding{55}                                     & \checkmark  & \ding{55}                                     & \checkmark                                      & \ding{55}                                      & \checkmark                                 \\ \hline
STSAN~\cite{li2020predicting} & \ding{55} & \ding{55}                                     & \checkmark  & \ding{55}                                     & \ding{55}                                     & \ding{55}                                      & \ding{55}                                \\ \hline
PMF~\cite{feng2020pmf}        & \checkmark  & \ding{55}                                     & \checkmark  & \ding{55}                                     & \checkmark                                      & \ding{55}                                      & \ding{55}                                \\ \hline
\citet{ezequiel2022federated}      & \ding{55} & \ding{55}                                     & \ding{55} & \ding{55}                                     & \ding{55}                                     & \ding{55}                                      & \ding{55}                                \\ \hline
STLPF~\cite{wang2022location} & \ding{55} & \ding{55}                                      & \checkmark  & \ding{55}                                     & \ding{55}                                     & \checkmark                                       & \ding{55}                                \\ \hline
\end{tabular}
\end{table}

\paragraph{\textbf{~\citet{fan2019decentralized}}} proposed a federated attention-based personalized human mobility prediction. They apply a \textit{few-shot} learning human mobility predictor that makes personalized predictions based on a few records for each user using an attention-based model. Furthermore, they take advantage of  pre-training strategies where the predictor is trained on another smaller mobility dataset   to accelerate the FL training on devices. However, even with the pre-training and attention-based strategy, the model requires over 1000 rounds of data communication rounds and is not sufficiently robust for  the irregular nature of the human movement.

\paragraph{\textbf{STSAN~\cite{li2020predicting}}}
\citeauthor{li2020predicting} in 2020  proposed  a  cross-silo personalized next point prediction model named STSAN (Spatial-Temporal Self-Attention Network)   which   integrates \textit{AMF} (Adaptive Model Fusion Federated Learning) for offering a  mixture of a local and global model.  The spatial attention layer allows for capturing the user’s preference for geographic location, and temporal attention  captures the user's temporal activity preference. To overcome the non-iid challenge, the AMF function enables the algorithm to learn specific personalization at each aggregation step on the FL server. The approach is evaluated on Foursquare, Twitter, and Yelp datasets.   Table~\ref{tab:stoa} reports its performance on the Foursquare dataset  against centralized prediction approaches   and shows the superior performance of 99\% APR and 6\% increase in Acc\@1 compared to VANext~\cite{gao2019predicting} .

\paragraph{\textbf{PMF~\cite{feng2020pmf}}}
~\citeauthor{feng2020pmf} proposed PMF, a privacy-preserving mobility prediction framework that uses FL to train general mobility models in a privacy-aware manner. In PMF, every participating device trains locally a representation of the global (centralized) model by using only the locally available dataset at each device. The framework also accounts for  attack cases in the mobility prediction task and uses a group optimization algorithm on mobile devices to tackle these attacks.  In the group optimization procedure, the whole model is divided into the risky group trained with protected data and the secure group trained with normal data. Furthermore,   an efficient aggregation strategy based on robust convergence and an effective polling schema for fair client selection in the centralized server. The results of this model on DenseGPS and Twitter dataset show similar top-1 performance to DeepMove~\cite{feng2018deepmove} but on Foursquare Tokyo it performs poorer than centralized baselines and other FL approaches.

\paragraph{\textbf{STLPF~\cite{wang2022location}}} 
~\citeauthor{wang2022location} proposed a spatial-temporal location  prediction  framework  (STLPF) where the next point prediction algorithm is trained based on a self-attention layer that enables information to be learned between long sequences  in both local and global models. Furthermore, as part of the framework, the authors propose an approach that enables   clients to cooperatively train their models in the absence of a global model. Their evaluation shows marginal improvement in the accuracy of next point prediction~(APR) on the Foursquare dataset when compared to the centralized approaches, and their approach performs similarly to STSAN~\cite{li2020predicting} (See Table~\ref{tab:acc}).
\vspace{0.2cm}

\paragraph{\textbf{\citeauthor{ezequiel2022federated}} ~\cite{ezequiel2022federated}}  develop two implementations of GRU-Spatial and Flashback on FL for predicting the next location in human mobility.  To the best of our knowledge, they are currently the only work that has worked on baselining these different approaches in an FL framework, namely Flower~\cite{beutel2020flower} and measuring the computational complexity of the model. They evaluate their model on  Foursquare NY and Gowalla datasets.

Building upon our exploration of trajectory predictions, it is noteworthy to acknowledge a distinct but closely related research direction—mobility mode prediction. In this realm, the focus shifts towards understanding and dissecting individuals' personal transportation modes, a facet that holds paramount importance in the efficient management of public transportation systems. Particularly in bustling metropolis cities marked by substantial populations and heavy traffic volumes~\cite{masek2016harmonized}. Decoding users' traffic modes and aggregating traffic patterns on a broader scale can profoundly influence decision-making processes. For instance, it provides valuable insights for adaptively allocating public transportation resources, and strategically assigning more support during peak times or on congested streets to alleviate traffic burdens.

Within this avenue, the work presented by~\cite{yu2022privacy} takes center stage. They propose a federated VGG-like model expressly designed for predicting users' transportation modes, utilizing the GeoLife dataset. This distinctive approach contributes to the evolving landscape of mobility mode prediction, offering insights that extend beyond trajectory predictions and bear significance in refining resource allocation strategies and bolstering the overall management of public transportation systems.~\citet{mensah2022efeddnn} extends this setting by considering three different architectures (GRU, LSTM and 1D CNN) trained simultaneously, obtaining three global
models after convergence. These global models' outputs are combined through an ensemble learning technique (\ie, stacking) with an MLP 
playing the role of the meta-learner. The objective of this MLP is to find the most confident label. (\ie, majority vote). The approach is 
tested against a federated version of each of the three architectures, showing clear improvement.

\subsubsection{Metrics}
Trajectory prediction models are commonly   benchmarked on a handful of available datasets using \textit{Acc@K} metric which is computed as an average of how many times the correct location was within the top-k predicted places (sorted by the model’s output weights). For example, for an Acc@5 metric, the target (or actual output) is compared against a vector of the top-5 most probable locations output by the model. If the target is an element of the top-5 vector, the prediction is correct (or true positive). Papers commonly report on Average Percentile Rank (APR) which is the average Percentile Rank for the prediction of a location. 

\subsubsection{Datasets}
The following  mobility datasets are used to evaluate the trajectory predictive approaches in the research community:

\begin{itemize}
\item 
\textbf{Foursquare~\cite{yang2014modeling}:} It comprises two sub-datasets, Tokyo and New York data. The Tokyo dataset contains 0.5 million check-ins in Tokyo while the New York one contains over 0.2 million, both collected over a span of about 10 months (from 12 April 2012 to 16 February 2013). Each check-in includes an anonymized user ID, timestamp, and location information, e.g., GPS coordinates and semantic meaning (represented by fine-grained venue categories).

\item 
\textbf{Twitter~\cite{zhang2016gmove} }: It contains around 1.1 million geo-tagged tweets from Los Angeles. These tweets are collected from 1 August 2014 to 30 November 2014. Every geo-tagged tweet consists of four parts, e.g., an anonymized user ID, location information (GPS coordinates), timestamp, and the message published by the user. Compared with the other two platforms, Twitter data is very sparse when location service is not a frequently-used function for Twitter users.

\item 
\textbf{Gowalla~\cite{gowalla}}: It consists of two parts: a check-in dataset and a friendship network dataset. The check-in dataset contains over 6.4 million check-ins contributed by more than 196,000 users, collected over the period of February 2009 to October 2010. Similarly to Foursquare, each check-in includes an anonymized user ID, latitude and longitude coordinates, a timestamp, and a location ID. As for the friendship network dataset, it contains information about social relationships between users, represented by over 95,000 undirected edges.

\item 
\textbf{Brightkite~\cite{gowalla}}: This dataset is similar to Gowalla in that it includes check-in data and a friendship network. However, this dataset is significantly sparser, as check-ins were deliberately shared by users, leading to a sparser dataset. Quantitatively, the dataset includes nearly 4.5 million check-ins and 58,228 users, collected between April 2008 and October 2010.

\item 
\textbf{Weeplaces~\cite{zhang2021double}}: This dataset has been sourced from Weeplaces, a website that provides visual representations of users' check-in activities on location-based social networks (LBSNs). The platform has been integrated with other location-based social networking services, such as Facebook Places, Foursquare, and Gowalla, through APIs. This dataset contains 7,658,368 check-ins from 15,799 users, as well as their friends present on Weeplaces.

\item 
\textbf{GeoLife~\cite{zheng2010geolife}:} It is a GPS trajectory dataset that was collected from 182 users over a span of five years (from April 2007 to August 2012). This dataset comprises 18,670 trajectories, each represented by a sequence of time-stamped points containing latitude, longitude, and altitude information. These trajectories capture a diverse range of users' outdoor movements, including routine activities (e.g., going home), as well as leisure activities (e.g., shopping, hiking, and cycling). Recently, 69 out of the 182 users have labeled their trajectories with transportation modes, such as driving, taking a bus, riding a bike, and walking. The labels for transportation mode are stored in a separate file for each user's folder.

\item 
\textbf{Yelp~\cite{liu2017experimental}:} It is a collection of businesses, reviews, and user data extracted from the Yelp platform. This dataset is regularly updated and contains almost 7 million reviews of over 150,000 businesses located in 11 metropolitan areas across the United States and Canada. The data includes information on individual users such as their name, the number and nature of their reviews, and their list of friends. Additionally, the dataset also includes check-ins, which provide information about the frequency and duration of customer visits to businesses.

\item
\textbf{Priva'Mov~\cite{mokhtar2017priva}: }It comprises data collected from multifarious sensors, including WIFI, GPS, and Cellular, and contains around 286.7 million records, where each record is a timestamped trajectory point containing latitude, longitude, userID. It was collected from 100 users over a period of 15 months and primarily focuses on urban mobility around Lyon city of France.

\item
\textbf{DenseGPS~\cite{feng2018deepmove}:} It is a dataset that includes  private data from a major mobile application provider in China with  5000 users with one-month dense location records. This dataset is not available for the research community to use.

\end{itemize}

\begin{figure}
\begin{subfigure}[b]{0.47\textwidth}
    \centering
    \includegraphics[width=\textwidth]{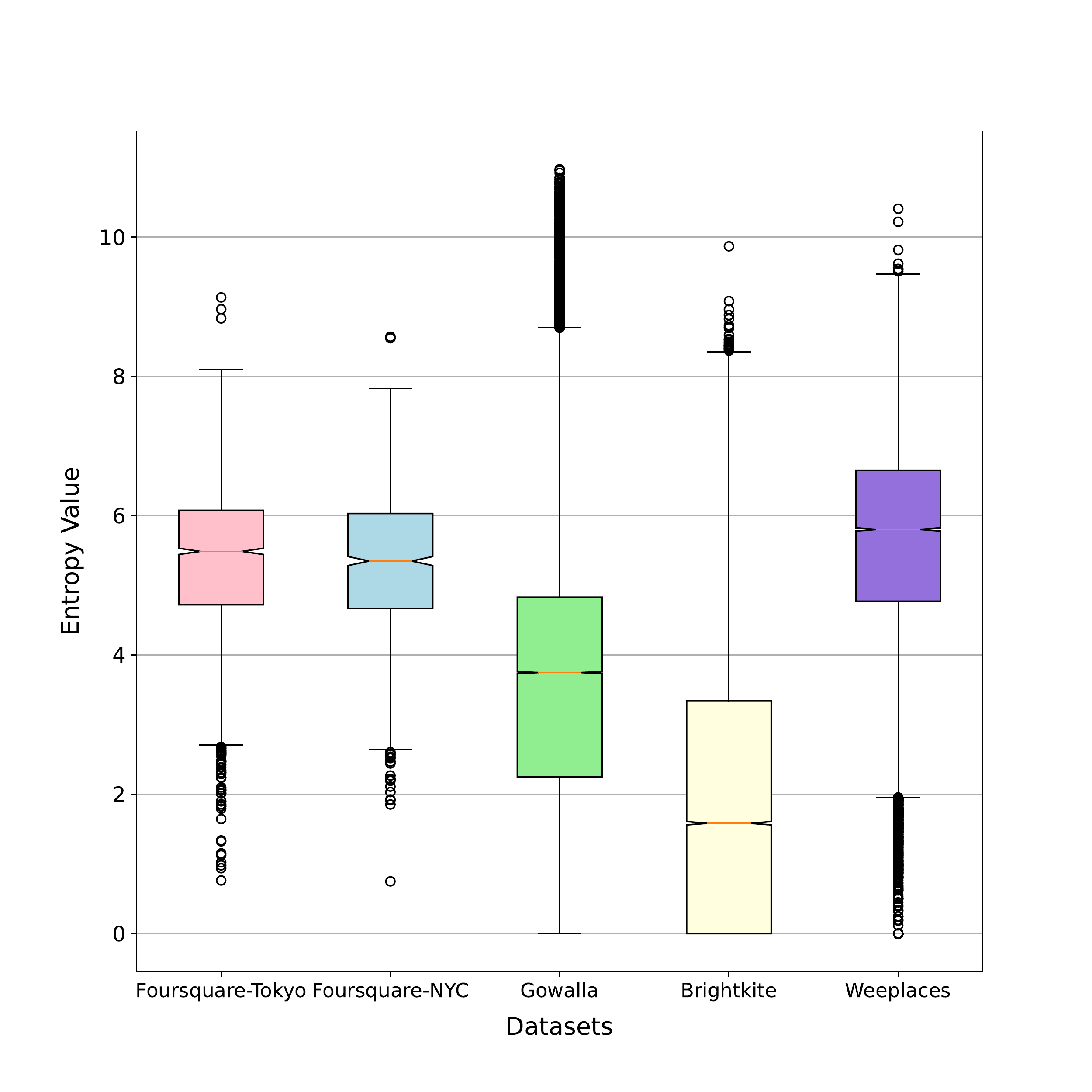}
    \caption{Horizontal entropy $H_u$ across different check-in datasets.}
    \label{subfig:entropycheckinperuser}
\end{subfigure}
\hfill
\centering
\begin{subfigure}[b]{0.47\textwidth}
    \centering
    \includegraphics[width=\textwidth]{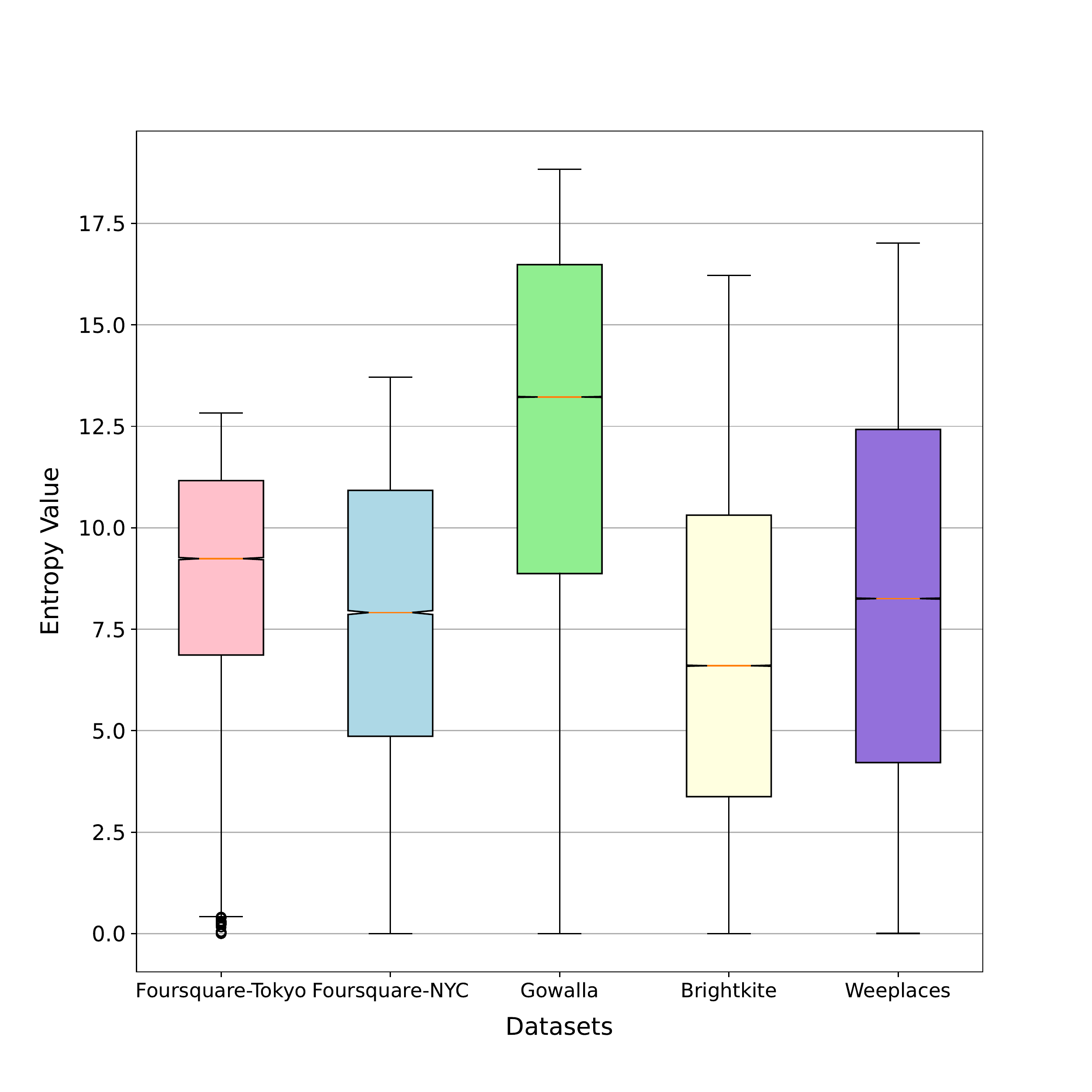}
    \caption{Vertical entropy $H_l$ across different check-in datasets.}
    \label{subfig:globalentropycheckin}
\end{subfigure}    
    
    \caption{Entropy Experiments.}
    \label{fig:entropycheckin}
\end{figure}

\begin{table}
\def\arraystretch{1.5}
\centering\captionsetup{justification = centering}
\caption{Global Statistics across check-in datasets.}
\label{tab:datasetstats}
\begin{tabular}{|c|c|c|c|c|c|c|}
\rowcolor{Gray}
\hline 
                 & $N_u$   & $N_l$    & $N_c$    & $\Tilde{N_u}$ & $\Tilde{N_l}$   \\ \hline
Foursquare-Tokyo & 2293   & 61858   & 573703  & 3.43 & 92.44  \\ \hline
Foursquare-NYC   & 1083   & 38333   & 227428  & 2.37 & 84.05  \\ \hline
Gowalla          & 107092 & 1280969 & 6442892 & 3.11 & 37.18  \\ \hline
Brightkite       & 50686  & 772780  & 4491080 & 1.39 & 21.16  \\ \hline
Weeplaces        & 15799  & 971308  & 7658368 & 2.72 & 166.64 \\ \hline
\end{tabular}
\def\arraystretch{1}
\end{table}


\subsubsection{Data sparsity and heterogeneity in trajectory prediction tasks}

 Mobility literature defines the highest potential accuracy of predictability of any individual, termed as ``maximum predictability'' ($\Pi_{max}$)~\cite{lu2013approaching}. Maximum predictability is defined by the entropy of information of a person's trajectory (frequency, sequence of location visits, etc.).  We adopt this measure to compute the non-iid property of the mobility traces. We use Shannon’s Entropy $H(x)$ to get a sense of both the sparsity and non-iidness of several check-in datasets. In this process, we adjust the metric proposed by~\cite{salem2022quantifying} to extend the individual entropy of users (i.e., horizontal) with an individual entropy of point-of-interests (i.e., vertical). We argue that having these two dimensions for the entropy is necessary to draw conclusions w.r.t to sparsity and heterogeneity. To quantify this relationship, we first measure the horizontal entropy in the following manner:  
\begin{equation}\label{eq:entropyhorizontal}
    H_u(x)  = - \sum_{i=1}^{n} P_u(x_i)\log_2[P_u(x_i)]
\end{equation}
Where $n$ is the number of POIs and is the size of the probability vector. $P_{u}(x_i)$ is the probability of an individual user $u$ visiting location $x_i$ considering exclusively spatial pattern.

We compute the vertical entropy as follows: 

\begin{equation}\label{eq:entropyvertical}
    H_l(x)  = - \sum_{u \in U}  P_u(l)\log_2[P_u(l)]
\end{equation}
Where $U$ is the set of all users. $P(x)$ is the probability of user $u$ visiting location $l$.

In Figure~\ref{fig:entropycheckin}, a comparison of entropy levels is shown for different check-in datasets. Meanwhile, Table~\ref{tab:datasetstats} provides a summary of more conventional statistics about these datasets. More specifically, $N_u$ denotes the number of users, $N_l$ denotes the number of POIs, $N_c$ denotes the number of check-ins, $\Tilde{N_u}$ denotes the average number of users that visited each point of interest and $\Tilde{N_l}$ denotes the average number of POIs visited by each user. Figure~\ref{fig:entropycheckin} seems to indicate that Brightkite has the least entropy, as expected. In fact, Subfigure~\ref{subfig:entropycheckinperuser} even shows that a significant proportion of Brightkite users have zero entropy, indicating that some users have only one check-in. This correlates with the high sparsity levels of the dataset (refer to Table~\ref{tab:datasetstats}). This downward trend is also evident when viewed through the lens of vertical entropy, although it may be less pronounced. Considering these observations, we see this dataset as an appropriate option to tackle the sparsity problem in distributed learning on mobility data. 
Conversely, Weeplaces exhibits two important trends: Firstly, there is a considerably higher level of horizontal entropy compared to what is seen on Brightkite, without a corresponding increase in vertical entropy. This suggests that in general users are more active and less predictable, but have not explored a much wider spectrum of points-of-interest. Secondly, Figure~\ref{subfig:entropycheckinperuser} exhibits a significant number of user outliers, particularly those with low entropy, indicating the presence of a short tail. These observations, correlated with the results of the table~\ref{tab:datasetstats}, which shows a high $N_l$ and a quite low $N_u$, indicate low heterogeneity levels within this dataset. Oppositely, Gowalla   exhibits more horizontal entropy, which indicates that users are more active globally while also having a higher vertical entropy, indicating that users have substantially less in common than in WeePlaces. These two observations suggest a more heterogeneous, yet less sparse dataset. This makes Gowalla a perfect candidate for works tackling heterogeneity. However, the presence of a significant proportion of highly mobile outlier users should also be noted. Finally, Foursquare datasets show good entropy levels on both axes, making them an excellent choice for sanity tests.

\subsection{Traffic Flow Prediction  Approaches}
Traffic Flow Prediction~(TFP) is a problem with multifarious applications. For instance, it allows to mitigate traffic, evaluate air pollution, estimate travel time, and improve driving experience. Approaches in predicting the traffic patterns range from parametric approaches, which are based on statistical metrics (\eg, ARIMA~\cite{}), to more advanced ML models such as those based on Recurrent Neural Networks~(RNNs) and Graph Neural Networks~(GNNs). For a comprehensive survey of deep learning models for traffic predictions in centralized settings refer to~\cite{jiang2021dl}. Flow prediction approaches require slightly different settings in FL than trajectory prediction. Unlike individual predictive approaches where each client is assumed to correspond to an individual with their mobility traces, in flow predictive approaches, the clients are often considered to be entities or organizations that are maintaining their flow data private to comply with privacy regulations. Considering this, Cross Silo FL appears as a promising paradigm for ensuring that records are not shared outside of each organization/sensor. Moreover, as the number of clients is smaller, and the datasets sizes per client is greater, this configuration alleviates some of the non-iidness concerns. However, this paradigm shift also ushers in two pivotal challenges: First, The nature of traffic flow prediction demands real-time adaptation due to the highly dynamic nature of the data. Consequently, the FL process cannot afford to wait until a significant data pool accumulates. Implementing online learning strategies becomes imperative to continuously update the model with incoming data streams. Second, the units responsible of collecting the data (\eg, sensors) often grapple with limited computational and spatial resources. Crafting efficient solutions becomes crucial to ensure the practicality of these solutions and their adaptability to real world applications. In the following, we categorize TFP works into two groups, depending on the types of models employed : grid-based approaches relying temporal models (\eg, RNNs, LSTM, etc.) and graph-based approaches employing GCNs. We first compare the literature on a high level before diving deeper into each work. We then introduce the metrics and datasets used for this task.

\subsubsection{\textbf{Grid based Approaches}}
In grid-based approaches, the input data into the model is a sequence of flow data per location over time (\eg, average speed, number of vehicles). These approaches mainly focus on deep neural models such as RNN, GRU and LSTM to capture past historical traffic information as a predictor for future instances. Table~\ref{tab:tfp-grid} provides a comprehensive summary of the challenges addressed by the current body of literature in this category. Notably, our assessment reveals a notable gap in addressing the resource limitations inherent in sensor networks, encompassing constraints related to memory and computational capabilities. Moreover, with the exception of two notable works~\cite{meese2022bfrt, yuan2022fedtse}, the overhead imposed by the proposed solutions remains largely unquantified. Furthermore, the current array of works largely overlooks the critical aspect of robustness—neglecting to account for potential vulnerabilities wherein sensors may transmit compromised model updates due to malicious interference or system faults.

It is also imperative to highlight that, apart from the noteworthy contribution by~\citet{sepasgozar2022fed}, there remains a dearth of comparative analyses with federated competitors beyond standalone evaluations—an aspect that deserves more attention in this field. A more comprehensive discussion of these shortcomings is expounded upon in Section~\ref{sec:futur}

\begin{table}[!ht]
\centering
\caption{Summary of the challenges tackled by the reviewed Grid-Based works.}
\label{tab:tfp-grid}
\begin{tabular}{rrrrrrrr}
\hline
                                                       & Privacy                    & \begin{tabular}[c] {@{}c@{}}Byzantine
                                                         \\Resilience
                                                         \end{tabular}                 & Non-IIDness                                                                           & \begin{tabular}[c]{@{}c@{}}Resource\\ constraints\end{tabular}            & \begin{tabular}[c]{@{}c@{}}Overhead \\ assessment\end{tabular}                & \begin{tabular}[c]{@{}c@{}}Federated\\ competitors\end{tabular} & \begin{tabular}[c]{@{}c@{}}Online\\ Learning\end{tabular} \\ \hline

FedGRU\cite{liu2020privacy}     & \ding{55}& \ding{55}&\checkmark                                                          & \ding{55}  & \ding{55}                                                                                                                    & \ding{55}                                     & \ding{55}                               \\ \hline
\citet{akallouch2022prediction} & \checkmark  & \ding{55}& \ding{55}                                                           & \ding{55}& \ding{55}                                                                                                      & \ding{55}                                     & \ding{55}                               \\ \hline

FedLSTM~\cite{tang2022differentially} & \checkmark&
\checkmark & \ding{55} & \ding{55} & \ding{55} & \ding{55} &
\ding{55} \\ \hline
BFRT\cite{meese2022bfrt}        & \ding{55}& \checkmark  & \ding{55}                                                           & \ding{55}& \checkmark                                                                                                              & \ding{55}                                     & \checkmark                                 \\ \hline
FedTSE\cite{yuan2022fedtse}     & \ding{55}& \ding{55}& \checkmark                     & \ding{55}& \checkmark & \ding{55}                                     & \checkmark                                 \\ \hline
Fed-NTP\cite{sepasgozar2022fed} & \ding{55}& \ding{55}& \ding{55}                                                           & \ding{55}& \ding{55}                                                         & \checkmark                                       & \ding{55}                               \\ \hline
\citet{zeng2021multi}           & \ding{55}& \ding{55}& \checkmark & \ding{55}& \ding{55}                                                                                  & \ding{55}                                     & \checkmark                                 \\ \hline
\end{tabular}
\end{table}

\paragraph{\textbf{FedGRU}~\cite{liu2020privacy}}
In their work, \citet{liu2020privacy} introduced a federated learning algorithm for highway flow prediction, leveraging a Gated Recurrent Unit (GRU). Additionally, they extended the client sampling phase of federated learning through a joint-announcement protocol. This protocol allows willing clients to announce their participation, strategically aiming to reduce communication costs associated with the FedAvg aggregation step. To address the challenges posed by non-IID data, especially in the context of highly heterogeneous locations, the authors suggested clustering clients based on their location information into $K$ clusters, utilizing a constrained K-means approach~\cite{bradley2000constrained}. This results in a model per cluster, and ensemble learning is then applied to identify the optimal subset of global models, enhancing prediction accuracy.

Their methodology was rigorously tested using real-world data from the Caltrans Performance Measurement System (PeMS)~\cite{chen2001freeway}, comprising 39,000 individual sensors monitoring the freeway system in real-time across major metropolitan areas of California. The results demonstrated comparable performance to the centralized baseline of the GRU model and various other centralized models.

\paragraph{\textbf{~\citet{akallouch2022prediction}}}
in this work, the authors address privacy concerns in the context of TFP. They tackle this issue by introducing Local Differential Privacy~(LDP) to provide theoretically provable privacy guarantees. Their approach involves training an LSTM model in a federated fashion, with clients transmitting noisy gradients where the noise is sampled from a Gaussian distribution. Despite the common expectation of a performance drop with such privacy-preserving mechanisms, their experimental results defy this trend, revealing competitive performance comparable to centralized baselines.

\paragraph{\textbf{FedLSTM}~\cite{tang2022differentially}} the authors strive to introduce both privacy and resilience guarantees into the context of TFP, operating under the distinctive assumption of computationally capable Roadside Units~(RSUs) alongside traditional Federated Learning (FL) clients, represented by individual vehicles. In this scenario, the authors devise a protocol wherein clients engage with RSUs during each FL round, transmitting their model updates. These RSUs, equipped with uniform test sets, evaluate and validate the models through consensus mechanisms involving miners. The validated models are subsequently published on the blockchain, with local aggregation by clients occurring at a later stage.

To enhance privacy, the authors adopt a strategy akin to~\cite{akallouch2022prediction}, leveraging Local Differential Privacy (LDP). However, they employ a distinct mechanism involving Laplacian noise, introduced to the count of vehicles in specific areas. The efficacy of this protocol is assessed using an LSTM and compared against various centralized baselines. Notably, the evaluation extends to scenarios with malicious users introducing poisoned models, showcasing the protocol's resilience due to the presence of verifiers. This work contributes a unique perspective by considering both privacy and resilience aspects in TFP.

\paragraph{\textbf{BFRT}~\cite{meese2022bfrt}} 
In this contribution, the authors also employ blockchain technology to enhance accountability and verification aspects within the Federated Learning (FL) framework. Specifically, they utilize a permissioned blockchain, implemented using Hyperledger Fabric~\cite{androulaki2018hyperledger}, to establish a framework where clients can only submit their model updates with permission from a designated group of verifiers (referred to as peers). Acceptance is contingent upon approval by a group of orderers through a consensus algorithm. While sharing a conceptual similarity with~\cite{tang2022differentially}, the distinguishing factor lies in the unique assumptions of this work. Here, the peers and orderers are presumed to be under the ownership of an institution, such as a service provider or government, with Roadside Units (RSUs) serving the role of clients in this setup.

Notably, the authors do not explicitly outline the criteria used by the verifiers (peers) to validate models. Evaluation of this protocol is conducted on a Delaware Department of Transportation (DelDOT) dataset, encompassing both GRU and LSTM architectures. A key distinction of this work, in contrast to previous ones~\cite{kaur2023federated,liu2020privacy,akallouch2022prediction,tang2022differentially}, \textit{is its focus on real-time TFP, even within the experimental settings, as opposed to offline learning scenarios. This real-time consideration adds a practical dimension to their exploration of FL in the context of TFP.}

\paragraph{\textbf{FedTSE}~\cite{yuan2022fedtse}}In FedTSE, authors proposed a framework for Travel State Estimation (TSE). They  design a long short-term memory (LSTM) model as the local training model for joint prediction of vehicular speed and traffic flow. A unique characteristic of FedTSE is that it relies on the deep reinforcement learning (DRL)-based algorithm to adjust model parameter uploading/downloading decisions  such that it  improves the estimation accuracy of local models and balances the tradeoff between computation and communication cost. They evaluate their approach on the England Freeway Dataset which includes flow and speed for the entire year of 2014.  They also consider three different aggregation strategies corresponding to a synchronous aggregation with the same number of epochs per client, namely, \textit{FedTSE-Syn}; an asynchronous aggregation with different numbers of epochs, namely, \textit{FedTSE-Asyn} and a weighted vrsion of FedTSE-Asyn where clients that have trained more are given more weight.

\begin{figure}
    \centering
    \includegraphics[width=0.5\textwidth]{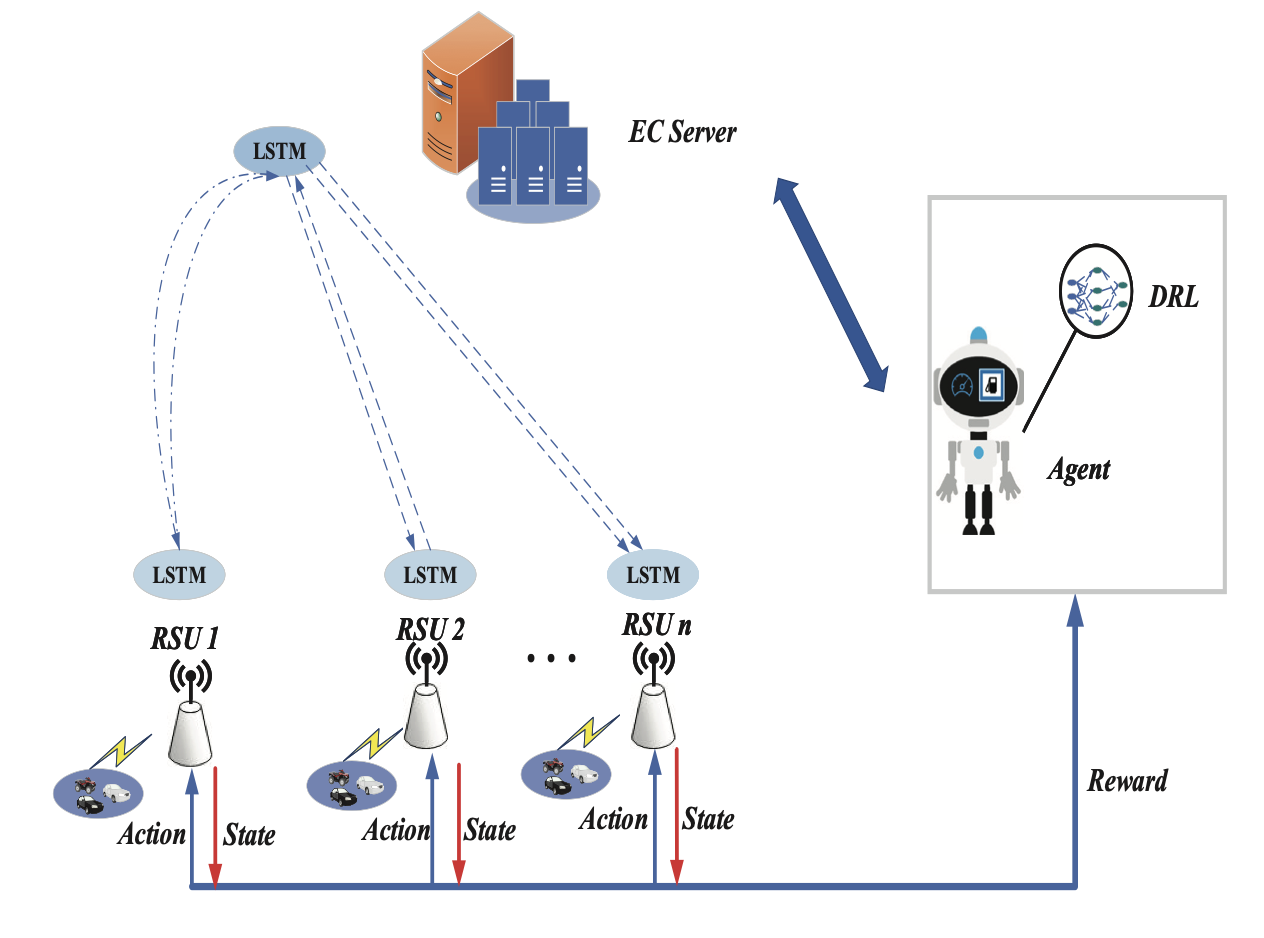}
    \caption{Figure from FedTSE~\cite{yuan2022fedtse} showing the interactions between Edge Computing (EC) Server and Road Side Units (RSU) acting as a cross-silo unit. The LSTM model weights are aggregated using FedAvg, and a Deep Reinforcement Learning (DRL) Agent to maximize reward.   }
    \label{fig:fedtse}
\end{figure}


\paragraph{\textbf{Fed-NTP}~\cite{sepasgozar2022fed}} proposed an LSTM model implemented in  FL to predict network traffic based on the most influential features of network traffic flow in   the Vehicular Ad-Hoc Network (VANET). Even though this work is \textit{not} aiming in predicting vehicle traffic and focuses on network flow, it has some relevance with the existing models such as FedGRU~\cite{liu2020privacy}. Fed-NTP shows to outperform FedGRU on the same V2V dataset. Other similar works that have focused on network traffic prediction such as FedDA~\cite{zhang2021dual}  exist in the literature.   However, as they solely focus on network traffic  problems and are not intended for mobility applications, we do not review them in the survey. We encourage the reader to refer to \cite{jiang2022cellular}  for a full survey of network traffic prediction techniques in FL. 

\paragraph{~\citet{zeng2021multi}}
To our knowledge, this work stands as the sole work introducing a multi-task learning method for the TFP challenge within the FL framework. The motivation behind this approach is to craft personalized models capable of adjusting to the diverse forms of heterogeneity previously outlined~(See Section\ref{sec:fl}). The authors' methodology begins with a hierarchical clustering of clients' (\ie, sensors) local data based on various criteria such as weather conditions, time, and special events. This clustering ensures the presence of consistent clusters across clients. Subsequently, an LSTM model is trained in a federated manner for each cluster to capture temporal features and address time travel prediction. This process yields a dynamic graph delineating sub-segments connecting different stations, along with their respective travel times—this graph evolves over time. Interestingly, the authors propose a modification of the $A^*$ algorithm to determine the optimal route utilizing this time-dependent graph. The experimental comparison with Google Maps' route prediction demonstrates the advantages of this approach. Nonetheless, the paper lacks explicit details on the clustering criteria employed and the mechanisms for client consensus regarding these clusters.
\subsubsection{\textbf{GNN Based Approaches}}
State-of-the-art multi-layer GNNs effectively address the spatial-temporal nature of traffic prediction in centralized settings. However, when applied in Federated Learning (FL) scenarios, GNN-based approaches encounter specific challenges. Notably, vertical FL poses a hurdle where each silo possesses a partial, overlapping view of the graph.

Motivated by intelligent transportation systems, such as Road Side Units (RSUs), individual units exhibit unique traffic patterns across a city (Figure~\ref{fig:fedtse}). The primary challenge in using GNN models for traffic forecasting within FL arises from their abundance of parameters, contrasting with simpler models like the Auto-Regressive Integrated Moving Average (ARIMA) with significantly fewer parameters (on the order of hundreds).Table~\ref{tab:params} shows the magnitude of parameters for baseline traffic prediction models over the PEMSD7M dataset~\cite{yu2017spatio}.

Moreover, the dynamic nature of these graphs over time demands adaptation through online learning, posing another significant challenge. Existing works aim to address these issues, as summarized in Table~\ref{tab:tfp-gnn}. While most studies tackle heterogeneity challenges, considerations regarding the cost of training and storing these complex models are rarely highlighted. Notably,~\cite{zhang2022gof} stands out by addressing these concerns, delegating the bulk of learning to a centralized server while clients handle and store manageable parameter subsets.

Furthermore, similar to grid-based approaches, the current literature lacks exploration into the models' robustness against faults and attacks that could compromise the learning process.

\begin{table}[!ht]
\caption{Number of parameters and performance of baseline centralized models on PEMSD7M~\cite{yu2017spatio} dataset.}
\label{tab:params}
\begin{tabular}{cccc}
  Model & Year & Number of Parameters & Performance (MAPE\%) \\
  \hline
  STGCN~\cite{yu2017spatio} & 2017 & 330K  & 5.02\% \\   
  MTGNN~\cite{wu2020connecting} & 2020 & 433K & 5.02\% \\
  DCRNN~\cite{li2017diffusion} & 2017 & 610K & 5.33\%  \\
\end{tabular}
\end{table}

\begin{table}[!ht]
\caption{Summary of the challenges tackled by the reviewed GNN-based works.}
\label{tab:tfp-gnn}
\begin{tabular}{rrrrrrrr}
\hline
                                                         & Privacy                    & \begin{tabular}[c] {@{}c@{}}Byzantine
                                                         \\Resilience
                                                         \end{tabular}
                                                         & Non-IIDness                & \begin{tabular}[c]{@{}c@{}}Resource\\ constraints\end{tabular}           & \begin{tabular}[c]{@{}c@{}}Overhead \\ assessment\end{tabular} & \begin{tabular}[c]{@{}c@{}}Federated\\ competitors\end{tabular} & \begin{tabular}[c]{@{}c@{}}Online\\ Learning\end{tabular} \\ \hline
Fed-STGRU\cite{kaur2023federated} & \ding{55} & \ding{55} & \checkmark  & \ding{55} & \ding{55}                                     & \ding{55}                                      & \ding{55}                                \\ \hline
FedAGCN\cite{qi2023fedagcn}       & \ding{55} & \ding{55} & \checkmark  & \ding{55} & \ding{55}                                     & \ding{55}                                      & \ding{55}                                \\ \hline
FedSTN\cite{yuan2022fedstn}       & \ding{55} & \ding{55} & \ding{55} & \ding{55} & \checkmark                                      & \checkmark                                       & \ding{55}                                \\ \hline
CTFL\cite{zhang2022graph}         & \ding{55} & \ding{55} & \checkmark  & \ding{55} & \checkmark                                      & \ding{55}                                      & \ding{55}                                \\ \hline
GOF-TTE\cite{zhang2022gof}        & \checkmark  & \ding{55} & \checkmark  & \checkmark & \ding{55}                                     & \checkmark                                       & \checkmark                              \\ \hline
\end{tabular}
\end{table}



\paragraph{\textbf{DST-GCN}~\cite{wang2022federated}} 
The authors parallel FedGRU~\cite{liu2020privacy} by adopting a GRU model to predict temporal dependencies  in TFP. However, they extend this approach by integrating a Graph Convolutional Network~(GCN) to capture spatial features, recognizing the GCN's aptitude for modeling inter-spatial characteristics. Augmenting this fusion, an attention mechanism is employed to identify and elevate critical spatio-temporal dependencies that dynamically evolve. Notably, the study promotes a client participation scheme based on both willingness and the evaluation of clients' models prior to aggregation. Their experimentation with the PeMS dataset substantiates that this client selection strategy significantly enhances performance compared to traditional FL.

The authors parallel FedGRU~\cite{liu2020privacy} by adopting a GRU model for traffic flow prediction. However, this model is only used to capture the temporal dependencies. To capture the spatial features, they introduce a novel approach by integrating the GRU with a GCN. By design, GCNs have a spatial character, allowing them to better model inter-spatial features. Authors also include an attention mechanism module in order to detect and give more importance to the critical spatio-temporal dependencies, which tend to change overtime. Finally, the authors propose to make the participation of clients based on each client's will, as much as their merit (\ie, clients' models are evaluated before aggregation). Authors evaluated their work on the PeMS dataset and concluded that their clients selection policy improves the performance over standard FL.

\paragraph{{\textbf{Fed-STGRU}~\cite{kaur2023federated}}}
Similar to~DST-GCN~\cite{wang2022federated}, this work propose a combination of a GRU and a GCN to solve TFP. Moreover, the authors also focus on improving the clients selection process of FL through a clustering strategy, where they opt for the Fuzzy c-means algorithm. This alternative is highlighted as less computationally demanding compared to the constrained k-means proposed in~\cite{liu2020privacy}.

In terms of performance, this work demonstrates superior results compared to a standalone FedAvg algorithm, showcasing the effectiveness of their hybrid GRU-GCN model and the efficiency of Fuzzy c-means for client clustering in the context of traffic flow prediction.

\paragraph{\textbf{FedAGCN}~\cite{qi2023fedagcn}}In this work, the authors propose a comprehensive solution to address the efficiency of Graph Convolutional Networks~(GCNs) in handling spatial correlations, a critical aspect of traffic flow data. Previous methods, particularly those not based on graph models, rely on individual clients learning on their local view of the system, leading to a neglect of spatial associations of the traffic network's topology. To overcome this limitation, the authors introduce GraphFed, a carefully devised algorithm aimed at enhancing communication efficiency while preserving the ability to capture spatial correlations. The key innovation lies in the division of the graph network into sub-graphs, with representative clients randomly selected for each sub-graph. These representatives aggregate data from all clients within their respective sub-graphs, facilitating the classification of the model's parameters into global and local categories, the latter being tailored specifically to spatial relationships.

Importantly, the practicality of GraphFed is evident as it mitigates the need to store the entire traffic network at the client level, a cumbersome requirement for large traffic networks. By operating on disjoint regions of the traffic network, representative clients efficiently avoid the necessity of sharing local parameters with the Federated Learning (FL) server, ensuring that only global parameters are learned in a federated manner, thereby minimizing communication overhead. The authors also adopt the ADGCN model~\cite{qi2021adgcn}, enhancing its efficiency by refining the convolution operation. This meticulous approach guarantees the adeptness of the model in addressing spatial correlations and overcoming challenges associated with federated learning on a broad scale within traffic networks.

\paragraph{\textbf{FedSTN}~\cite{yuan2022fedstn}}
The FedSTN approach as formalized by Yuan et. al is  a newly proposed solution to solving the TFP problem.  To accomplish this task, authors proposed a Graph-based Representation Learning for CS FL using three main modules: a  Recurrent Long-term capture Network (RLCN) module, an Attentive Mechanism Federated Network (AMFN) Module, and a Semantic Capture Network module (SCN). The RLCN module is responsible for learning long-term traffic behaviors as geometric time series data of fixed long-term interval $p$. The data as input into this module is comprised of inflow-outflow values for certain grid spaces, as initially computed for the data. 

The AFMN module is responsible for learning short-term spatial-temporal features in a privacy-preserving manner. This module includes long-term contextual data such as meteorology  and federated graph attention (FedGAT).    Lastly, there is the SCN module, which takes into account Point of Interest Feature Components (POI) and non-euclidean connection relationships. Points of interest and their interactions have significant effects on TFP but are not incorporated into the raw time series prediction data. Furthermore, transportation networks also have methods of connection outside adjacent-grid connections, such as trains, highways, etc. These flows also have an important effect on TFP that is not incorporated elsewhere, so this module is to address this issue.  The output of all these modules is then connected via an FC layer followed by a Tanh activation function for the final output.

\paragraph{\textbf{CNFGNN}~\cite{meng2021cross}} \citeauthor{meng2021cross} proposed a new   architecture named Cross Node Federated Graph Neural Network (CNFGNN) to predict the flow.   CNFGNN works by decomposing the problem into two stages: first,  it uses an encoder-decoder network to  extract temporal features locally, and a  then GNN to capture spatial relations across devices. On each training step, the server processes a temporal encoding update, a partial gradient update, and an on-node graph embedding update.  Each iteration updates the client side weights and then ships the model, hidden layer, and gradients back to the server.

\begin{figure}
    \centering
    \includegraphics[width=0.45\textwidth]{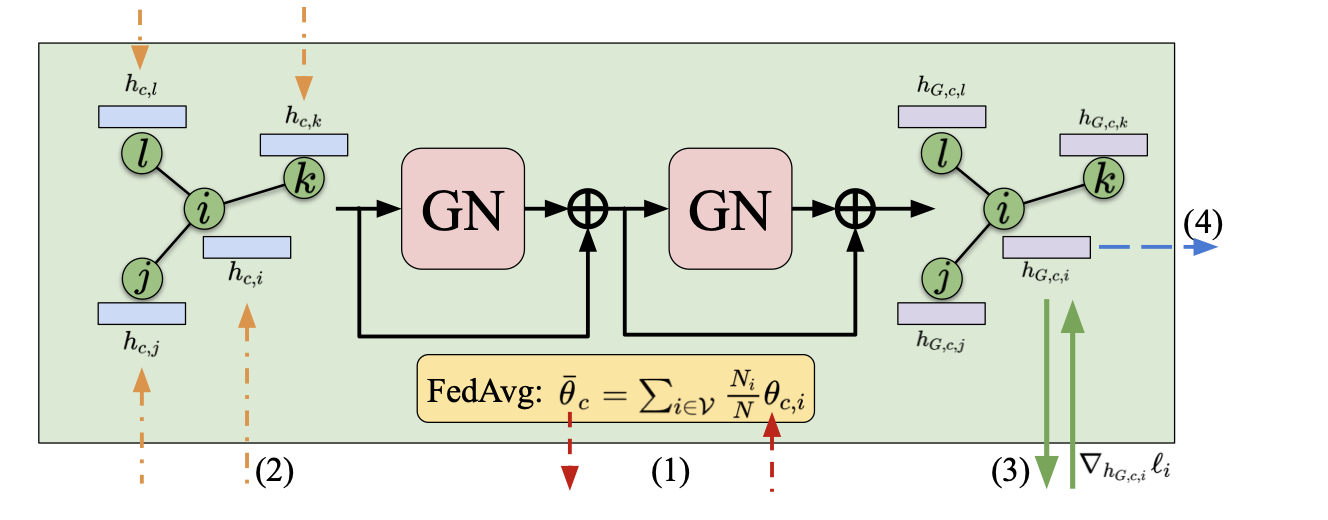}
    \includegraphics[width=0.45\textwidth]{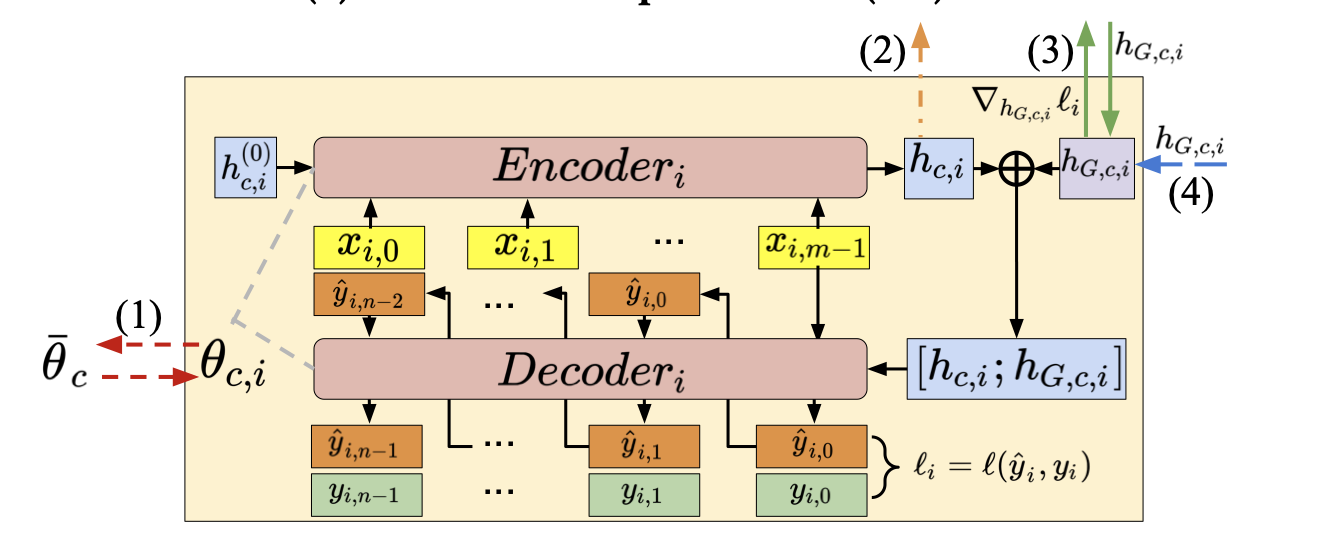}

    \caption{Figure from CNGNN~\cite{meng2021cross} presents (a) the server-side graph neural network with a systematic overview of training steps: (1) Federated learning of on-node models. (2) Temporal encoding update. (3) Split Learning of GN.  (b) Client $i$ Auto-encoder architecture.}
    \label{fig:cngnn}
\end{figure}
One large issue of the proposed approach is the communication overhead in the training stage, with split learning requiring the global model to fetch all hidden states from each node, and ship gradients of node embeddings to each node. Then, it must receive gradients of node embedding and send gradients of hidden states in the back-propagation step. To mitigate this, they propose an alternative training approach in which the temporal encoder-decoder and the node embeddings are trained separately. First, the node embeddings are fixed and optimization is performed on the encoder-decoder. Then, after a fixed interval, the global model is updated fixing the node-level models. This drastically reduces communication overhead. 
The FL local models and the GNN model with only a local objective function. They perform alternating optimization to update clients’ model weights with GNN model weights fixed and then update GNN model weights with the FL local model weights fixed, over multiple rounds.

\paragraph{\textbf{CTFL}~\cite{zhang2022graph}} proposes a Clustering-based hierarchical and Two-step- optimized FL (CTFL), to   overcome the large number of parameters that are needed for aggregation in the   GNN-based models such as STGCN~\cite{yu2017spatio}, DCRNN~\cite{li2017diffusion} and MTGNN~\cite{wu2020connecting}. CTFL employs a divide-and-conquer strategy, clustering clients based on the closeness of their local model parameters. It also accounts for optimization by applying a  two-step strategy where the central server  uploads only one representative local model update from each cluster, thus reducing the communication overhead associated with model update transmission in the FL. 

\paragraph{\textbf{GOF-TTE}~\cite{zhang2022gof}} This work addresses a different formulation of the TFP problem, namely, Time Travel Estimation~(TTE), where the task is to predict the travel time needed to gom from a point $a$ to a point $b$ considering spatial features (\eg, road network map) and temporal features (\eg, time of the day). ~\cite{zhang2022gof} tackles this challenge within the context of taxi-driving scenarios, with potential applications to diverse driving types. In this work, generic spatio-temporal features are used to learn the global state of the network in a federated manner, allowing to obtain a global model. Subsequently, each client fine-tunes this model in locally~(\ie, Localized global model), before incorporating a personalized model trained on the clients' profile features.

The input architecture involves a dual graph representing road segments and intersections. These elements are translated into latent representations (\ie, embeddings) before being past to a GCN layer, allowing for the capture of spatial dependencies. Spatial and temporal representations are then subjected to a cross-product operation. Temporal aspects, such as day of the week and time of the day, are encoded, with an attention mechanism considering both current and past temporal impacts. This cross product results in a global state of the road network.

In a subsequent phase, the global state is fine-tuned using clients' local data before being obfuscated with Laplacian noise through the DP mechanism. A personalized model is later introduced, comprising a fully connected layer of client profile feature embeddings (\eg, frequently visited regions, average driving distance). This personalized model is integrated with the output of the localized global model to mitigate biases arising from the high heterogeneity of the data.

Notably, this work distinguishes itself from prior research by acknowledging the relationship between road segments and intersections, in contrast to road-segment-based solutions for instance. Furthermore, it introduces a novel personalization technique that accounts for the current global state of the network, enabling real-time personalized predictions—a combination of considerations not concurrently addressed in existing literature.

Finally,  it is worth noting that other works similar to flow prediction exists which are focused on identifying travel modality inference (TMI). For instance \cite{zhang2021toward} proposes MTSSFL which trains a deep neural network ensemble under a novel semisupervised FL framework. It  achieves a highly accurate  score for a  crowdsourced TMI without depending on the availability of massive labeled data.

\subsubsection{Metrics} The forecasting performance for traffic flow predictions is commonly measured as a mean absolute error (MAE), mean square error (MSE), root MSE (RMSE), and mean absolute percentage error (MAPE).

\subsubsection{Datasets}

For flow predictive approaches the following datasets are  widely used to benchmark the comparison of various algorithms.

\begin{itemize}
\item \textbf{Taxi Datasets:} The following datasets contain taxi in-out flow data collected via GPS, which include pick-up time, drop-off time, and trip distance.

\begin{itemize}
    \item \textbf{TaxiBJ~\cite{zhang2017deep}} contains  data collected in Beijing from four time intervals: 2013/7/1 to 2013/10/30, 2014/3/1 to 2014/6/30, 2015/3/1 to 2015/6/30, and 2015/11/1 to 2016/4/10. The data was collected at 30-minute intervals and include trajectories of over 34,000 taxis. It additionally contains meteorology data such as weather conditions, temperature, and wind speed.
    \item \textbf{TaxiNYC~\cite{taxinyc2009}} contains data collected in New York City, which was collected between 2016/4/1 and 2016/6/30 and contains over 35 million records. A more extensive version of this dataset, collected from 2009 to 2022, is also available.  
    \item \textbf{TaxiPorto~\cite{pkdd-15-predict-taxi-service-trajectory-i}} comprises a full year 
    (from 2013/7/1 to 2014/6/30) of trajectories for all the 442 taxis running in the city of Porto, Portugal. It also contains information on the type of taxi call: central-based, stand-based, and demanded on a random street.
    \item  \textbf{T-Drive~\cite{zheng2011t-drive}} contains GPS traces collected in Beijing, China. The dataset was collected over a period of three years from 2008 to 2011, and it consists of over 10,000 taxi drivers' GPS trajectories. The dataset contains a total of 1.07 billion GPS points, covering approximately 150 million kilometers. That being said, the sample released by Microsoft is over a span of one week only, containing around 15 million GPS points and covering a total distance of trajectories that reaches 9 million kilometers.
\end{itemize}

\item \textbf{METR-LA~\cite{jagadish2014big}: } is Los Angeles traffic collected using 207 sensors mounted around highways, and 1515 edges from 2012/3/1 to 2012/6/30. 
\item  \textbf{PeMS} is a traffic flow dataset collected from California Transportation Agencies Performance Measurement System (PeMS).

\begin{itemize}
  \item   \textbf{PeMS-BAY~\cite{li2017diffusion}: } It contains 325 nodes (traffic sensors) and 2369 edges in the Bay Area from 2017/1/1 to 2017/5/31.
\item \textbf{PeMSD7M: } It is a sub-sample of PeMS published as part of \cite{yu2017spatio}, also collected from PeMS. It covers 228 traffic sensors with a 5-minute sampling rate corresponding to  2012/5/1 to 2012/6/30.
\end{itemize}

\item \textbf{NY-Bike~\cite{nycbike2013}: } Spanning a period of ten years, from June 2013 to January 2023, this dataset includes comprehensive information about daily bike orders by people in New York City and is regularly updated. More specifically, it contains information about bike trips, such as duration, starting and ending point, and location, as well as details about bikers, including user type, gender, and year of birth.

\item \textbf{Yelp~\cite{liu2017experimental}:} It is a collection of businesses, reviews, and user data extracted from the Yelp platform. This dataset is regularly updated and contains almost 7 million reviews of over 150,000 businesses located in 11 metropolitan areas across the United States and Canada. The data includes information on individual users such as their name, the number and nature of their reviews, and their list of friends. Additionally, the dataset also includes check-ins, which provide information about the frequency and duration of customer visits to businesses.

\end{itemize}


%

\begin{table}[!ht]
\label{ref:tabFlow}
\caption{RMSE metric for reviewed Federated Learning models and centralized baselines over   various benchmark datasets.}
\begin{tabular}{llllllll} 
                                                     & \textit{NY Bike }     & \textit{Taxi-NY }   & \textit{Yelp}       & \textit{Taxi BJ }      & \begin{tabular}[c]{@{}l@{}}\textit{PeMS}\\ \end{tabular} & \textit{PEMS-BAY (5mn)}                                             & \textit{METR-LA (5mn)}                                              \\
\hline
\multicolumn{8}{c}{Centralized Approaches }                                        \\
\hline
ARIMA                                                & 10.07    & 12.43  & -  &  22.78  &  - &  5.59 & 7.66 \\
ST-RESNET~\cite{zhang2017deep}                                           &   6.33    & 9.67   &   -&  16.69  & - & - &- \\
\begin{tabular}[c]{@{}l@{}}GRU\\ GRU+NN\end{tabular} & - & - & - & - & 9.97   & \begin{tabular}[c]{@{}l@{}}4.12\\ 3.81\end{tabular} & \begin{tabular}[c]{@{}l@{}}11.78\\ 11.47\end{tabular} \\
\hline
\multicolumn{8}{c}{Federated Approaches }                                        \\
\hline
FedSTN~\cite{yuan2022fedstn}                                               & - & 9.32   & - & 24.22  & - & -&   - \\
FedGRU~\cite{liu2020privacy}                                               & 17.14   & -  & 1.22   & -  & 11.04    & - &   -   \\
Federated-LSTM~\cite{taik2020electrical}                                             & 17.24   & - & 1.24   & - &    -  &  -  & -  \\
MVFF (GRU+GNN)~\cite{errounda2022mobility}                                       & 6.79   & -  & 0.96   &  -  &  -  & - &  - \\
CNFGNN~\cite{meng2021cross}                                               & -           & -          & -          & -          & -                                                   & 3.82                                                 & 11.48                                                 \\                      
\end{tabular}
\end{table}

\subsection{Top-N Location-Based Recommendation Approaches}
Location-based recommendation task, while resembling trajectory prediction, often entail lower demands. This task involve identifying a top-K set of relevant Points of Interest (POIs) for a user based on their historical POIs. In Table~\ref{tab:recsys}, we summarize the challenges addressed by the community for this use case. Notably, none of the reviewed works have considered an online learning setting. One might argue that this task may not necessarily require online learning because users' preferences might not significantly change over time, unlike the trajectory prediction task, which inherently encompasses a natural time dimension. However, we argue that exploring online learning could still be valuable. Moreover, recent research has demonstrated that users' preferences tend to evolve over time. While other challenges have been reasonably explored, it's noteworthy that several works have based their privacy solutions on not sharing all model parameters, but recent studies on similar models have shown this approach not to be entirely private~\cite{zhang2023comprehensive,yuan2023interaction}. This is indicated in the table using the "$\approx$" symbol. Finally, similar to previous approaches/tasks, the exploration of robustness has been lacking. This is unfortunate, given that issues like shilling attacks~\cite{huang2023single} are significant concerns for recommendation systems and represent just one type of robustness concern not explored in this federated setting. Other types include malicious users fabricating fake profiles to manipulate recommendations, promote specific POIs, and introduce unfairness.

Given that the evaluation metrics and datasets align with those used in trajectory prediction approaches, the rest of this section focuses solely on the main works in this domain.

\begin{table}[]
\caption{Summary of the challenges tackled by the reviewed Top-N POI recommendation works.}\label{tab:recsys}
\begin{tabular}{rrrrrrrr}
\hline
                                                                                    & Privacy                                        & \begin{tabular}[c] {@{}c@{}}Byzantine
                                                         \\Resilience
                                                         \end{tabular}                                  & Non-IIDness                                    & \begin{tabular}[c]{@{}l@{}}Resource\\ constraints\end{tabular} & \begin{tabular}[c]{@{}c@{}}Overhead \\ assessment\end{tabular} & \begin{tabular}[c]{@{}c@{}}Federated\\ competitors\end{tabular} & \begin{tabular}[c]{@{}c@{}}Online\\ Learning\end{tabular} \\ \hline
FPL~\cite{anelli2021put}                                      & $\approx$                         & \ding{55}                    & \ding{55}                   & \checkmark                                      & \ding{55}                                   & \checkmark                                       & \ding{55}                              \\ \hline
DFL-PC~\cite{guo2022deep}                                           &  \ding{55}                                               &   \ding{55}                                              &  \checkmark                                              & \ding{55}                                                                &    \ding{55}                                                            &    \ding{55}                                                             &     \ding{55}                                                      \\ \hline
PREFER~\cite{guo2021prefer}                                   & $\approx$                         & \ding{55}                    & \ding{55}                   & \checkmark                                      & \checkmark                                      & \checkmark                                       & \ding{55}                              \\ \hline
MVFF~\cite{errounda2022mobility}                              &    $\approx$                                            &  \ding{55}                                               &   \checkmark                                             & \ding{55}                                                               &      \ding{55}                                                          & \ding{55}                                                                &      \ding{55}                                                     \\ \hline
PriRec\cite{chen2020practical}               & \checkmark                      & \ding{55}                    & \checkmark                      & \checkmark                                      & \ding{55}                                   & \ding{55}                                    & \ding{55}                              \\ \hline
PEPPER\cite{belal2022pepper}                                 & \ding{55}                   & \ding{55}                    & \checkmark                      & \ding{55}                                   & \checkmark                                      & \checkmark                                       & \ding{55}                              \\ \hline

FedPOIRec\cite{perifanis2023fedpoirec}                                & \checkmark                  & \ding{55}                    & \checkmark                      & \ding{55}                                   & \checkmark                                      & \checkmark                                       & \ding{55}                              \\ \hline

\end{tabular}
\end{table}

\paragraph{\textbf{FPL}~\cite{anelli2021put}} This work extends the bayesian pair wise ranking~(BPR)~\cite{rendle2012bpr} algorithm to the federated setting. The authors train the sensitive user embeddings locally and provide users with the option to share these embeddings with a certain probability, while training the less sensitive parameters in a federated manner. This approach can be viewed as a flexible framework for factorization models, where clients can decide how many parameters to share while maintaining convergence of the model. They evaluate their approach on Foursquare dataset, considering different countries and levels of sparsity and compare themselves with a federated movies recommender~\cite{ammad2019federated}. 

\paragraph{\textbf{DFL-PC}~\cite{guo2022deep}} This work focuses on optimizing the aggregation process in federated learning. The authors pre-train deep models consisting of a GCN and a GRU connected to a multilayer perceptron~(MLP), which are used to estimate the system parameter space. These estimates are then optimized at the server level through a reinforcement learning algorithm. The evaluation results show that their solution surpasses traditional centralized deep models due to the added optimization step based on reinforcement learning.

In terms of system architecture, alternative solutions have been proposed for more efficient federated POI recommendation. The following works fall within this scope: 
 
\paragraph{\textbf{PREFER}~\cite{guo2021prefer}} ~\citeauthor{guo2021prefer} proposed a two-set recommendation model training. First, for privacy purposes, they propose to train the user-dependent model parameters strictly locally. Subsequently, user share and aggregate less sensitive parameters (i.e., user-independent) in a multiple-edge server architecture, instead of remote cloud servers, with an aim to improve real-time response capability and reduce communication cost. They validate their approach both analytically and empirically on two standard POI recommendation models, namely, Distance2Pre~\cite{cui2019distance2pre} and PRME-G~\cite{feng2015personalized}, and two check-in datasets, Foursquare~\cite{yang2014modeling} and Gowalla~\cite{gowalla}, and show the competitiveness of their approach with centralized and federated competitors.

\paragraph{\textbf{MVFF}~\cite{errounda2022mobility}} proposes a vertical federated learning framework for mobility data forecasting  for CS FL applications where each organization holds a partial subset of data.  Using a local learning model, each organization extracts the embedded spatio-temporal correlation between its locations. To account for global learning, a global model synchronizes with the local models to incorporate the correlation between all the organizations’ locations.

\paragraph{\textbf{PriRec}~\cite{chen2020practical}} proposes a peer-to-peer approach to learn sensitive user embeddings, while less sensitive ones(\eg, feature interaction model) are learned in a federated manner. This is achieved through the introduction of secret sharing~\cite{shamir1979share} in the decentralized gradient descent (DGD) topology and considering geographical information when building this topology, that is, users closely geo-located learn (privately) collaboratively their respective user embeddings. As for the items' features, they are aggregated through a using aggregation protocol~(SA)~\cite{jahani2023swiftagg+}. Authors evaluate their approach on standard POI datasets' and compare it with a centralized factorization machine~(FM), which it seems to compete with, while guaranteeing a better level of privacy.

\paragraph{\textbf{PEPPER}}~\cite{belal2022pepper} emphasizes personalization of POI models through the aggregation step while using a gossip communication protocol to eliminate the central FL server. In this work, nodes gossip their models with their neighbors and aggregate them after evaluating their contribution. The authors also introduce a peer-sampling protocol that acts as a clustering over time, ensuring each node has similar users in its neighborhood. Results from their experiments show that fully decentralized federated learning can be competitive with centralized solutions, while offering scalability and personalization.

\paragraph{\textbf{FedPOIRec}}~\cite{perifanis2023fedpoirec} introduces a framework that leverages social relationships between users to make more personalized models. The authors first train a global model using federated learning and multi-party computation (SMPC) to protect the aggregation phase from a curious server. A trusted third party is then tasked with finding similar users based on encrypted embeddings for personalized aggregation. The encryption is done using a leveled variant of the CKKS fully homomorphic encryption (FHE) scheme. This works also encompasses an adaptation of the setting of~\cite{anelli2021put} and CESAR~\cite{tang2018personalized}, a sequential recommender based on a convolutional neural network. The authors validate their approach through an evaluation on five versions (cities) of Foursquare as well as a formal privacy analysis.

Finally, another use case that has been explored in the context of location-based recommendation systems is the driver recommendation use case, as tackled by~\cite{vyas2020vehicular}. In this work, cab companies use federated learning to strengthen roadside units (RSUs) with the computational capability to develop an intelligent recommender system that recommends the appropriate driver for a subsequent trip. To this end, they consider both the driver's stress and past behavioral patterns. \\

\subsection{Other Approaches}
\subsubsection{Clustering Based Approaches}

Most existing works in clustering in federated learning are focused on methods to  identify and self-organize devices  into communities are so to    conduct model sharing within those communities.  In~\cite{briggs2020federated}, authors introduce and evaluate a hierarchical clustering for vision models, where the local model is a Convolutional Neural Network (CNN) that is trained under \textit{supervised} learning.  The extensive evaluation presents the improvement that hierarchical clustering can bring to federated learning under a non-iid setting where each client holds partitioned data.  In~\cite{kim2021dynamic}, the authors propose the dynamic GAN-based clustering in FL to improve the time series forecasting for cases such as cell tower handover prediction. Their proposed approach accounts for the adaptive clusters and non-iid data. IFCA proposed by~\cite{ghosh2020efficient}   starts by randomly initializing $k$ models, one per cluster. Each client assigns itself to a cluster at the start of each round of training by evaluating all k models on its local data and choosing the model with the lowest loss to train for $m$ epochs. At the end of each round, the server   performs federated averaging within each cluster of clients separately. 

Although this theoretical line of work is receiving a great deal of attention from the Federated Learning research community we have seen almost no adaptation, except one, to the spatial-temporal models in FL. F-DEC~\cite{mashhadi2021deep} proposed a deep embedded clustering for urban community detection in federated learning. They expanded on the centralized model proposed by~\cite{ferreira_deep_2020} and trained an autoencoder based on  heatmap images of mobility trajectories transformed using the frequency of visits (where brighter pixels show more frequently visited areas). They then used a KL divergence loss for clustering similar heatmaps together. Furthermore, this work is  the only early evidence that we found that measures  the computational complexity of  such algorithms when it is deployed on ordinary smartphones.


\subsubsection{Privacy and Attacks in Spatial-Temporal FL Models}
Federated learning was initially designed to protect user privacy by sharing model parameters instead of data. However, research has shown that sharing these parameters can still reveal sensitive information, especially in models that use embeddings/latent features to capture user or point-of-interest semantics. To address this issue, researchers have proposed mainly three approaches: i) "a share less" policy, ii) injecting noise using techniques like Differential Privacy (DP), and iii) using cryptographic methods like Secure Multi-Party Computation (SMPC). Various studies have proposed solutions using each of these approaches, but each approach has its limitations.
For instance, ~\cite{guo2021prefer,anahideh2020fair} proposed sharing only user-independent embeddings to be learned in a federated manner while training user embeddings locally.~\cite{feng2020pmf} first proposed a practical attack, demonstrating that user check-ins can be easily inferred by a curious FL server based on POIs embeddings. Later on, they proposed to alleviate this attack by training these embeddings on noisy data generated using DP. They show that if the rest of the network (i.e., non-sensitive layers) is frozen during this noisy training, and is pre-trained on real data, then the performance remains reasonable. Nevertheless, they did not quantify the impact of their attack nor the degree of protection provided by their solution. Another line of work, ~\cite{perifanis2023fedpoirec}, opts for the third approach and uses SMPC to hide the individual contributions of the users from a curious FL server. Unfortunately, this approach opens the door to malicious users, whose goal would be to corrupt the learning, and who would be difficult to detect due to SMPC. 

A more privacy-oriented solution was proposed by ~\cite{chen2020practical}. They aggregated less sensitive embeddings using SMPC and categorized sensitive embeddings into two parts: those related to POIs and those related to users. They used Local DP to add noise at the user level to the POI-related embeddings before sharing them with the server. They also proposed to share the user-dependent embeddings in a peer-to-peer fashion using secret sharing. To reduce overhead and the attack surface, they considered geographic information to build the peer neighborhood. Thus, a user shares its sensitive embeddings only with geographically close neighbors. This solution has competitive prediction performances with non-private solutions and undeniably provides more privacy guarantees. However, the impact of the assumption that sensitive embeddings can be safely shared with nearby users is still unclear. \\

Another type of attack that has received significant attention in recent mobility research is re-identification attacks~\cite{maouche2017ap, gambs2014anonymization}. The fundamental concept behind such attacks is that a malicious service provider could exploit background knowledge to associate anonymized user traces with their respective owners, thereby compromising users' anonymity. To address this issue,~\citeauthor{khalfoun2021eden} proposed a federated protocol for assessing the risk of re-identification on mobility data. This protocol involves training a re-identification model in a federated manner using users' traces and subsequently utilizing this model to select the optimal combination and hyperparameters of location privacy protection mechanisms (LPPMs) that can protect a user's privacy whenever they transmit their data to an untrusted Mobile Crowd Sensing Server (MCS). Notably, this solution appears to be the only privacy risk assessment mechanism for mobility data that does not necessitate the presence of a trusted curator, to the best of our knowledge. For a full survey of privacy and security techniques in Federated Learning see~\cite{mothukuri2021survey}.

\section{Discussion and open research challenges}\label{sec:futur}

Based on the review of the above papers, we see the following opportunities and roadmap for the research community to explore. 

\subsection{Semantic Location Embedding and Context-Awareness Modelling}
\label{subsec:semantic}
One of the biggest opportunities that we see in continuing research on location and point trajectory predicting, is in regards to integrating more semantic and contextual information about types of places instead of focusing primarily on coordinates. 
For example, this contextual information can include information on whether a point in trajectory represents someone's workplace, their frequently visited locations, or a potential point of interest in a new town. 
While this is not limited to FL applications and previous works have often used open source maps to infer information about the type of places (e.g., popularity, socio-economic level), FL can bring a new level of anonymity and personalization to this integration. The semantic representation of the locations  can then be learned over time on users' devices, maintaining users' privacy while allowing for better-personalized models.

\subsection{Byzantine Resilient Spatio-Temporal Mobility Federated Learning}
\label{subsec:byzantine}
Within the realm of ML, Byzantine Resilience commonly denotes the ability to train an accurate statistical model amid the presence of arbitrary behaviors, commonly referred to as Byzantine users~\cite{guerraoui2023byzantine}. These behaviors manifest either due to faults or malicious users, encompassing scenarios like a compromised sensor in the TFP problem or a user aiming to promote specific items in a Top-N POI recommendation use case. In the best-case scenario, the trained model proves unusable, resulting in a cost loss incurred during its training. However, in the worst-case scenario, if the attack goes undetected (e.g., backdoors on GNN~\cite{zhang2021backdoor}), the model might exhibit concealed yet malicious behaviors. This predicament poses a significant challenge owing to the difficulty of interpreting model parameters.

Distinguishing between a Byzantine model update transmitted by a client or sensor and an honest but out-of-distribution client often presents a non-trivial task. This complexity amplifies in highly heterogeneous configurations typical of ST mobility FL applications. Its pertinence highlights this challenge as a substantial research track for Distributed ML and FL. Surprisingly, this aspect remains relatively unexplored in the context of ST mobility FL. Among the multitude of reviewed papers, only~\citet{meese2022bfrt} and~\citet{tang2022differentially} have delved into such adversarial settings. However, neither of these works has considered the novel solutions proposed in this field, such as bucketing~\cite{karimireddy2020byzantine} and variance reduction techniques~\cite{farhadkhani2022byzantine}. This gap is more concerning given the pronounced heterogeneity characteristic of these applications.

Consequently, there is an urgent imperative to delve into this challenge and contemplate environments accommodating Byzantine behaviors within FL clients. This pursuit should be seen as a fundamental step to bridge the gap between research efforts and real world implementations. Specifically, it necessitates a comprehensive study and evaluation of existing Byzantine-resilient distributed learning algorithms, a deep understanding of their limitations, and subsequently adapting them to suit the inherent heterogeneity and dynamic evolution within the spatio-temporal context.

\subsection{Communication Efficiency}
\label{subsec:communication}
The race to enhance the performance of machine learning (ML) models is driving exponential growth in their number of parameters across various fields. In the realm of ST mobility models, where the necessity to capture diverse dimensions (spatial, temporal, preferential, and characteristic) is paramount, this growth becomes even more pronounced. A striking illustration of this trend is the substantial increase of over 4000\% observed between ARIMA, a parametric model, and recent GNNs, which now number in the hundreds of thousands (see \cref{tab:params}).

Traditionally, in classical ML, concerns centered around the costs associated with training these models and the extensive data required for such endeavors. However, the emergence of FL has shifted the focus towards apprehensions about the expenses related to communicating and transmitting these increasingly expansive models. Surprisingly, a notable observation is that the majority of existing works do not adequately consider these challenges within their frameworks. That is, they do not consider both the cost of training as well as the cost of collecting these models by the central server.

Recognizing the significance of this issue, particularly in light of its comprehensive exploration in other fields, we advocate for heightened attention from the research community. We urge future work to delve into existing solutions like pruning~\cite{jiang2022model} and quantization~\cite{lang2023joint}, emphasizing the need to quantify the cost-effectiveness of these approaches in the context of ST mobility FL.

\subsection{Trust, Fairness, and Accountability} In addition to trust and accountability, another challenge that we see spatial temporal mobility models will face under a federated setting is fairness. \textit{That is to what extent the models that are trained on location traces are equitable?} Especially models that are designed for the purpose of mobility flow prediction and allocation  of transport options. For instance,   mobility demand prediction algorithms have been shown to offer higher service quality to neighbourhoods with more white people~\cite{brown2018ridehail}. Indeed, as recent evidence from the broader machine learning domain has shown, the systematic discrimination in making decisions against different groups has been shifted from people to autonomous algorithms~\cite{kasy2021fairness,heidari2019moral}. In many applications, discrimination may be defined by different protected attributes, such as race, gender, ethnicity, and religion, that directly prevent favourable outcomes for a minority group in societal resource allocation, education equality, employment opportunity, etc~\cite{sattigeri2019fairness}. Measuring fairness of mobility models   is a dimension that has been vastly overlooked in applications of spatial-temporal mobility models, with exception of a few works~\cite{yan2019fairst,yan2020fairness,ge2016racial} and with little consensuses on how fairness should be defined and measured for spatial-temporal applications. One way of controlling for fairness of mobility models under the FL setting is to create auditing systems that can infer information about  the training without having access to location data of the devices or the global model at the FL server~\cite{mashhadi2022auditing,mashhadi2022fairness}.   We believe future work will focus on dynamic middle-wares that can leverage solutions such as clustered FL  to offer interpretability of the underlying models~\cite{9443958,ghosh2020efficient,li2023towards} are crucial to transition exiting solutions from research to practice.

\subsection{Standardisation and Reproducibility}
\label{subsec:standardization}
As the landscape of spatio-temporal mobility research continuously evolves, it is increasingly crucial to establish a common ground for assessing its advancements. This involves not only pinpointing the most effective proposed solutions but also identifying the persistent challenges and shortcomings. Achieving this necessitates standardized datasets, uniform methods for data preprocessing and splitting, and ideally, a code base or metadata ensuring the reproducibility of each solution. Regrettably, similarly to other domains like recommendation systems~\cite{ferrari2021troubling,ferrari2019we}, this practice is seldom followed.

For instance, in studies related to the TFP task, there is a prevalent use of diverse datasets, sometimes from the same source but across disparate timeframes, different space and time discretizing methods as well as unclear distinct splitting methods. These splits often rely on various sampling techniques. Yet, as underlined in~\cite{meng2020exploring}, the process of partitioning data into training, testing, and potentially validation sets significantly influences the measured performance. Another example, specifically in Top-N POI recommendation, involves the utilization of metrics based on parameters (\eg, varying K values for accuracy at rank K). In FL, this challenge exacerbates due to the proliferation of parameters impacting the results (\eg, the number of clients, sampling strategies, aggregation techniques).

Consequently, robustly comparing different works, even those evaluated on identical datasets, becomes impossible. This is likely why most reviewed works unfortunately lack substantial comparisons with their federated counterparts. Moreover, considering that the distinct experimental procedures often stem from stochastic processes (client and data point sampling, stochastic learning), we stress the necessity for statistical tests to gauge the statistical significance of observed results.

We acknowledge that research contexts rarely align perfectly. Nevertheless, 
the need for standardizing datasets and preprocessing methodologies in the field is still undeniable. Existing surveys~\cite{zhao2016urban, freire2016exploring} have proposed solutions in a centralized context. While the direct applicability of these solutions to federated learning may require adaptation, they provide valuable insights and methodologies that can serve as foundational building blocks for the FL paradigm. Moreover, there have been several software efforts~\cite{yu2022transbigdata,haidri2022ptrail,graser2019movingpandas,pappalardo2019scikit,olive2019traffic,boeing2017osmnx,axhausen2007definition} that furnish a set of libraries for standardized spatio-temporal data analysis, preprocessing, and visualization. Using these as building blocks might represent an essential step towards 
standardization and reproducibility in the field.

\subsection{Realistic Cross-silo Spatial Temporal Datasets for Benchmarking}
Existing approaches that we reviewed are mostly evaluated with ST data partitioned artificially. 
Nevertheless, the long-term development of this field still requires realistic and large-scale federated datasets to be made available to support experimental evaluations under settings close to practical applications. 
For instance, in the reviewed literature, there is a lack of research on how the geographic distribution of silos can lead to a geographically distinct flow of information. 
Establishing policy-based scenarios in order to guide how the data should be partitioned across silos to reflect real-world data ownership challenges is a direction that we believe the research community will be  working closely with other stakeholders in the future.

\subsection{Transition to Real-world Deployment Through Dedicated Frameworks}
Finally, we believe that just as crowd-sensing research was successful a decade ago through frameworks such as AWARE~\cite{ferreira2015aware}, which reduced the burden of app development for data collection, frameworks specifically designed for federated mobility models will facilitate the transition from limited research to in-wild deployments. To achieve this transition, it is crucial to i) provide benchmarks for mobility applications and ii) develop mobility-centered federated learning frameworks, as was the case for graph applications~\cite{he2021fedgraphnn} and IoT applications~\cite{zhang2020federated}. This will allow the research community to effectively evaluate and compare the performance of federated learning models on mobility data.
We foresee that the transition between the current research efforts to real deployment will happen over stages where first multi-disciplinary research will focus on understanding users' attitudes towards using their location data for training models. After all,  similar research on crowd-sensing applications has shown that location information is a top concern of users' involvement in these applications ~\cite{data-usage-privacy-uk-consumers}.  To the best of our knowledge, currently, there are  no existing works in understanding users' privacy concerns when their data is not shared externally but is still used in creating predictive models. As a next step, we envision a slow transition between fully centralized models to decentralized models. Rather than training models focused on end-user prediction tasks, generative models that allow synthetic trace generation by learning from user mobility traces will be used to update and de-bias centralized datasets.

\section{Conclusion}\label{sec:conclusion}

In this paper, we  surveyed  the federated  learning models in the domain of mobility prediction  as well as the widely used datasets for spatial-temporal models. We described the challenges that exist in applying  common deep learning techniques in decentralized settings and discussed the opportunities for the research community to consider for future work. Our work indicates rapid growth and interest in this space, with promising future directions both in terms of theoretical frameworks and models, and practical applications and use cases. We hope both academics and practitioners find this survey useful for choosing the appropriate approach for their individual scenarios.
\newpage
\bibliographystyle{ACM-Reference-Format}
\bibliography{ref,aaairefs}

\section{Appendix}

\subsection{Survey Review Methodology}
To create the summary of the surveys and their topics, we crawled the citation (bib file) of all the published articles   with the keywords `Federated Learning' and `Survey' or `Review' in the title for each year.  We used the publish or perish tool for this crawling and used Google Scholar as the platform.  We downloaded the abstracts of all these surveys from those papers and fed them to a Large Language Model (GPT 3.5) for thematic categorization of the topics. We followed a similar process for counting the number of papers per year that were published on ST FL topics (non-surveys).

\subsection{Supplementary Tables}
\begin{table}[!h]
\centering
\caption{Federated learning definitions.}
\label{tab:fl_definitions}
    \begin{tabular}{ll}
    Variable & Description \\
    \midrule
    $G^t$ & global model at round $t$. \\
    $n$ &   total number of participants. \\
    $m$ &  subset of participants selected for a single round. \\
    $\eta$ & global learning rate. \\
    \textit{L} & locally trained model. \\
    $\mathcal{D}$ & local data. \\
    \textit{E} & number of epochs for local training. \\
    \textit{lr} & local learning rate.\\
     $S$ & clipping bound. \\
     $\sigma$ & amount of added noise. \\
\end{tabular} 
\end{table}

\begin{table}[!h]
\caption{RMSE metric for reviewed Federated Learning models and centralized baselines over   various benchmark datasets.}
\begin{tabular}{llllllll} 
                                                     & \textit{NY Bike }     & \textit{Taxi-NY }   & \textit{Yelp}       & \textit{Taxi BJ }      & \begin{tabular}[c]{@{}l@{}}\textit{PeMS}\\ \end{tabular} & \textit{PEMS-BAY (5mn)}                                             & \textit{METR-LA (5mn)}                                              \\
\hline
\multicolumn{8}{c}{Centralized Approaches }                                        \\
\hline
ARIMA                                                & 10.07    & 12.43  & -  & 22.78  &  - &  5.59 & 7.66 \\
ST-RESNET~\cite{zhang2017deep}                                           &   6.33    & 9.67   &   - & 16.69  & - & - &- \\
\begin{tabular}[c]{@{}l@{}}GRU\\ GRU+NN\end{tabular} & - & - & - & - & 9.97   & \begin{tabular}[c]{@{}l@{}}4.12\\ 3.81\end{tabular} & \begin{tabular}[c]{@{}l@{}}11.78\\ 11.47\end{tabular} \\
\hline
\multicolumn{8}{c}{Federated Approaches }                                        \\
\hline
FedSTN~\cite{yuan2022fedstn}                                               & - & 9.32   & - & 24.22  & - & -&   - \\
FedGRU~\cite{liu2020privacy}                                               & 17.14   & -  & 1.22   & -  & 11.04    & - &   -   \\
Federated-LSTM~\cite{taik2020electrical}                                             & 17.24   & - & 1.24   & - &    -  &  -  & -  \\
MVFF (GRU+GNN)~\cite{errounda2022mobility}                                       & 6.79   & -  & 0.96   &  -  &  -  & - &  - \\
CNFGNN~\cite{meng2021cross}                                               & -           & -          & -          & -          & -                                                   & 3.82                                                 & 11.48                                                 \\                      
\end{tabular}
\end{table}

\end{document}